\documentclass[lettersize,journal]{IEEEtran}
\usepackage{amsmath,amsfonts}
\usepackage{algorithmic}
\usepackage{algorithm}
\usepackage{array}
\usepackage{textcomp}
\usepackage{stfloats}
\usepackage{url}
\usepackage{verbatim}
\usepackage{graphicx}
\usepackage{cite}
\usepackage{multirow}
\usepackage{subfigure}
\usepackage{bbding}
\usepackage{makecell}
\usepackage{color}
\usepackage[colorlinks,
            linkcolor=blue,
            anchorcolor=blue,
            citecolor=blue]{hyperref}
\usepackage{orcidlink}

\begin{document}

\title{Clothes-Changing Person Re-Identification with Feasibility-Aware Intermediary Matching}

\author{Jiahe Zhao\textsuperscript{\orcidlink{https://orcid.org/0000-0003-1362-7673}}, Ruibing Hou\textsuperscript{$\dagger$ \orcidlink{https://orcid.org/0000-0003-2480-6538}}, Hong Chang\textsuperscript{\orcidlink{https://orcid.org/0000-0002-2668-0070}},~\IEEEmembership{Member, IEEE}, Xinqian Gu\textsuperscript{\orcidlink{https://orcid.org/0000-0003-1234-8795}}, \\Bingpeng Ma\textsuperscript{\orcidlink{https://orcid.org/0000-0001-8984-205X}},~\IEEEmembership{Member, IEEE}, Shiguang Shan\textsuperscript{\orcidlink{https://orcid.org/0000-0002-8348-392X}},~\IEEEmembership{Fellow, IEEE} and Xilin Chen\textsuperscript{\orcidlink{https://orcid.org/0000-0003-3024-4404}},~\IEEEmembership{Fellow, IEEE}
\thanks{$\dagger$ Corresponding author.}
\thanks{Jiahe Zhao, Hong Chang, Xinqian Gu, Shiguang Shan and Xilin Chen are with Key Laboratory of Intelligent Information Processing, Institute of Computing Technology (ICT), Chinese Academy of Sciences (CAS), Beijing, 100190, China, and University of Chinese Academy of Sciences, Beijing, 100049, China. (e-mail: \{zhaojiahe22s, changhong, sgshan, xlchen\}@ict.ac.cn, xinqian.gu@vipl.ict.ac.cn)}
\thanks{Ruibing Hou is with Key Laboratory of Intelligent Information Processing, Institute of Computing Technology (ICT), Chinese Academy of Sciences (CAS), Beijing, 100190, China. (e-mail: houruibing@ict.ac.cn)}
\thanks{Bingpeng Ma is with the School of Computer Science and Technology, University of Chinese Academy of Sciences, Beijing, 100049, China. (e-mail: bpma@ucas.ac.cn)}}

\markboth{Journal of \LaTeX\ Class Files,~Vol.~14, No.~8, August~2021}%
{Shell \MakeLowercase{\textit{et al.}}: A Sample Article Using IEEEtran.cls for IEEE Journals}


\maketitle

\begin{abstract}
Current clothes-changing person re-identification (re-id) approaches usually perform retrieval based on clothes-irrelevant features, while neglecting the potential of clothes-relevant features. However, we observe that relying solely on clothes-irrelevant features for clothes-changing re-id is limited, since they often lack adequate identity information and suffer from large intra-class variations. On the contrary, clothes-relevant features can be used to discover same-clothes intermediaries that possess informative identity clues. Based on this observation, we propose a Feasibility-Aware Intermediary Matching (FAIM) framework to additionally utilize \textbf{clothes-relevant features} for retrieval. Firstly, an Intermediary Matching (IM) module is designed to perform an intermediary-assisted matching process. This process involves using clothes-relevant features to find informative intermediates, and then using clothes-irrelevant features of these intermediates to complete the matching. Secondly, in order to reduce the negative effect of low-quality intermediaries, an Intermediary-Based Feasibility Weighting (IBFW) module is designed to evaluate the feasibility of intermediary matching process by assessing the quality of intermediaries. Extensive experiments demonstrate that our method outperforms state-of-the-art methods on several widely-used clothes-changing re-id benchmarks. 
\end{abstract}


\section{Introduction}
\label{sec: introduction}
\IEEEPARstart{P}{erson} re-identification (re-id)~\cite{gu2019temporal,hou2020iaunet,sun2018beyond} aims at identifying the target person across different times and locations captured by surveillance systems. Most existing works~\cite{bai2022salient,gu2020appearance,hou2021bicnet,wang2018learning,gu2023msinet,zhou2023adaptive,ji2022asymmetric} assume that a person always wears the same clothes, which may not be applicable in situations where individuals change their clothes. However, clothes changing is commonplace in long-term re-id scenarios. Due to its crucial role in intelligent surveillance systems, clothes-changing re-id has received increasing attention.

Recently, clothes-changing re-id works~\cite{gu2022clothes,hong2021fine,yang2023good,han2023clothing} focus on extracting clothes-irrelevant features to remove the reliance on clothing appearances. A line of work resorts to auxiliary modalities, \textit{e.g.}, skeletons~\cite{qian2020long}, silhouettes~\cite{jin2022cloth, cui2023dcr} and 3D shape~\cite{chen2021learning, liu2023learning}, to capture identity information that remains independent of clothes. Another line of work attempts to extract clothes-irrelevant features based solely on RGB modality. This is achieved through various techniques like adversarial learning~\cite{gu2022clothes}, semantic feature augmentation~\cite{han2023clothing}, causal intervention~\cite{yang2023good}, and association-forgetting learning~\cite{liu2023clothes}.

\begin{figure}[tbp]
    \centering
    \includegraphics[width=1.0\linewidth]{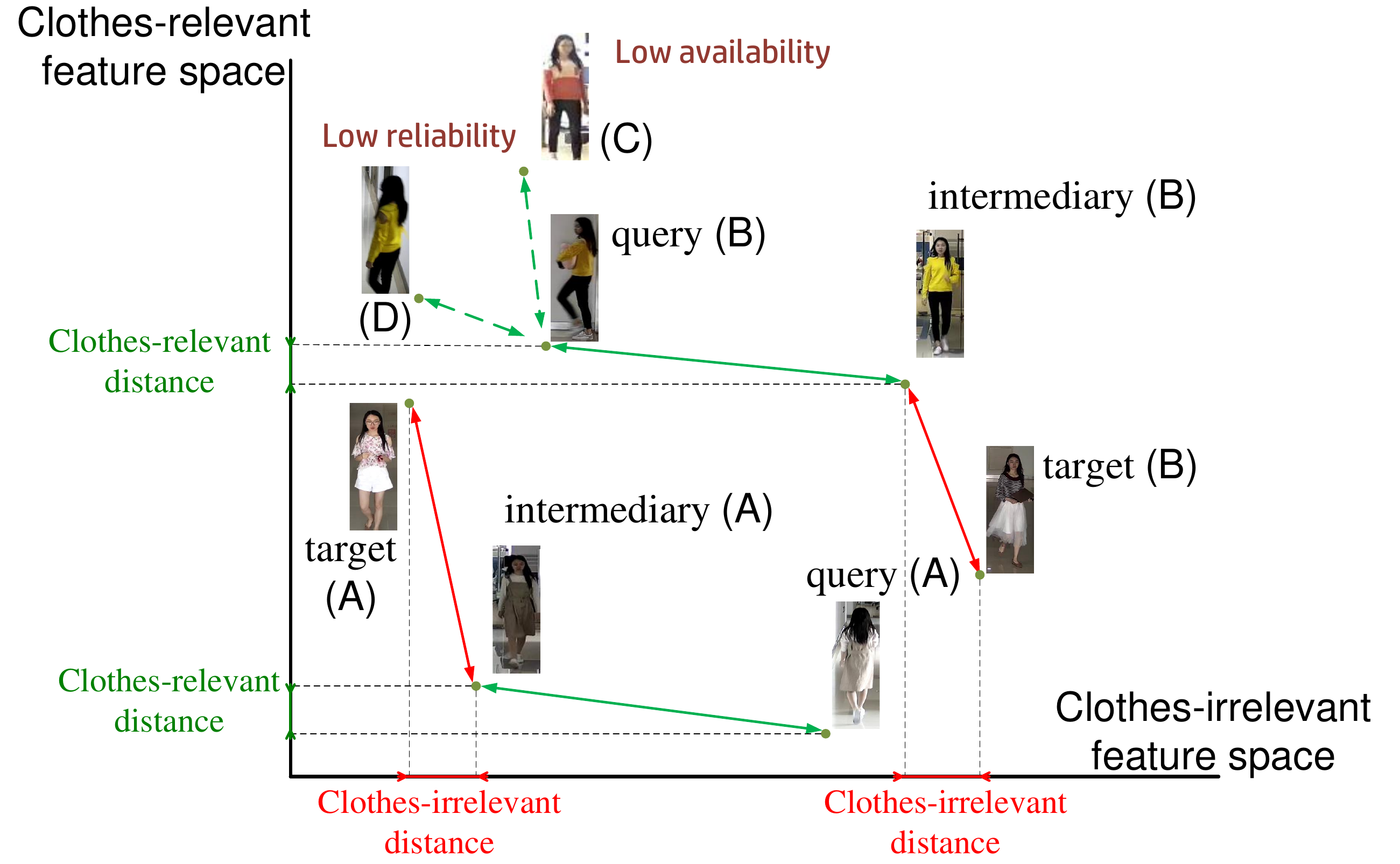}
    \caption{Illustrative examples of intermediary matching approach. (A) for query lacking clothes-irrelevant identity information (facial representation), we can match it to target through an intermediary with clear facial information. (B) for the query and target suffering large intra-class variation (body shape), we can match them through an intermediary with aligned body shape. (C) and (D) represent intermediaries of query (B) with low availability and low reliability, respectively. }
    \label{fig: idea illustration}
    \vspace{-3mm}
\end{figure}

As evident from current research, in clothes-changing scenarios, the person is represented solely by clothes-irrelevant features. However, clothing-irrelevant features may fail to provide sufficient information due to two challenges: (1) \textit{Inadequate identity information in clothes-irrelevant characteristic}. As pedestrian images are naturally taken on various camera views, the face characteristic could often be obscured or entirely invisible, while the body shape characteristic could be incomplete, as illustrated in the query ($A$) image in  Fig.~\ref{fig: idea illustration}. (2) \textit{Large intra-class variations of clothes-irrelevant characteristic}. Due to large variations in person pose and scale, clothes-irrelevant characteristics, such as body shape and sketch, could vary significantly within a person, as illustrated in the query ($B$) and target ($B$) images in Fig.~\ref{fig: idea illustration}. Therefore, it is necessary to explore richer clues beyond clothes-irrelevant characteristics for clothes-changing re-id.

In order to address the above challenges, we propose to utilize clothes-relevant features which are often overlooked in clothes-changing scenarios. Though clothes traits might fail to discriminate between different identities that wear similar clothes, we can circumvent this problem by learning clothes-relevant features that contain identity-discriminative information. To supervise the clothes-relevant features, we use specially designed clothes label, where each identity category is further divided into multiple fine-grained categories of clothes. This design ensures that samples with different identities never share the same label, even if they wear same or similar clothes. Supervised by this specially designed clothes label, our clothes-relevant features are capable of discriminating between identities, which suits well to the clothes-changing setting. Moreover, the clothes-relevant features offer additional advantages over their clothes-irrelevant counterparts. Specifically, clothes-relevant features convey identity information through the clothes appearance. This identity information remains visible and consistent under pose and view changes, ensuring information adequacy and intra-class invariance. Consequently, when a sample lacks clothes-irrelevant clues, by alternatively leveraging clothes-relevant features, we can fetch samples with the same clothes and abundant clothes-irrelevant clues. These samples can then serve as intermediaries for matching clothes-changing targets.

\begin{figure}[tbp]
    \centering
    \begin{subfigure} 
        \centering
        \includegraphics[width=0.85\linewidth]{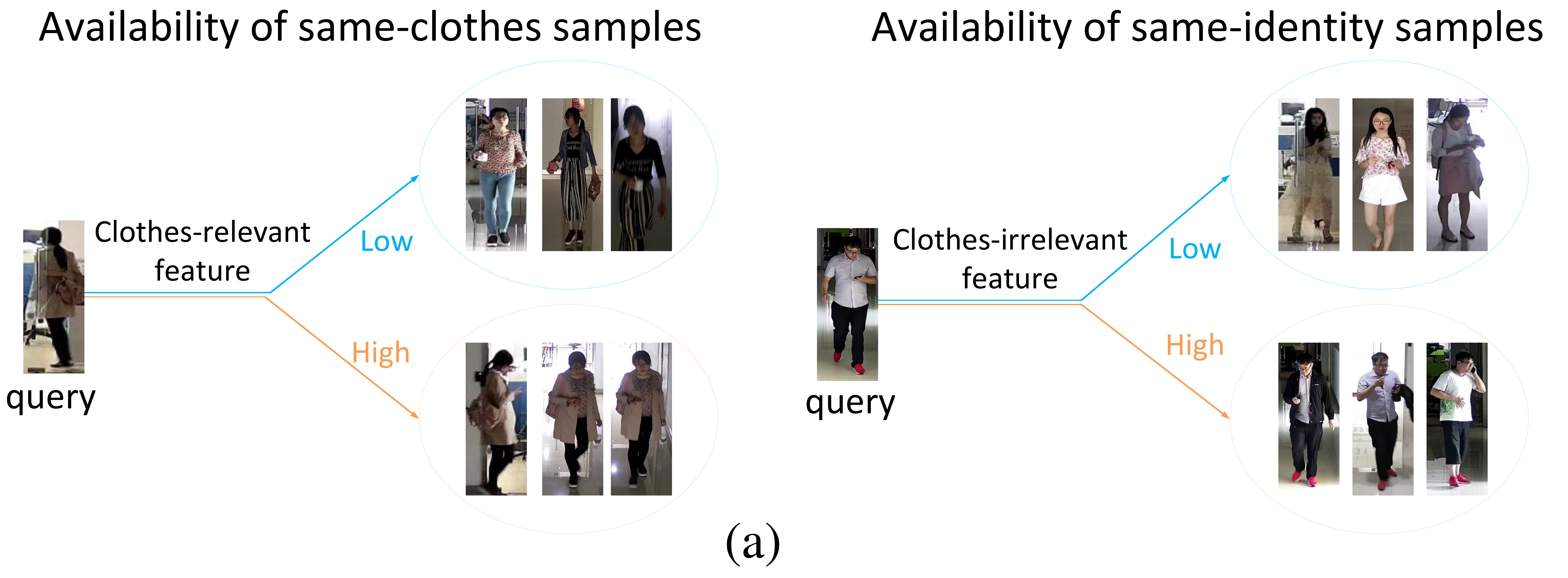}
        \label{fig:availability examples}
    \end{subfigure}
    \begin{subfigure}
        \centering
        \includegraphics[width=0.85\linewidth]{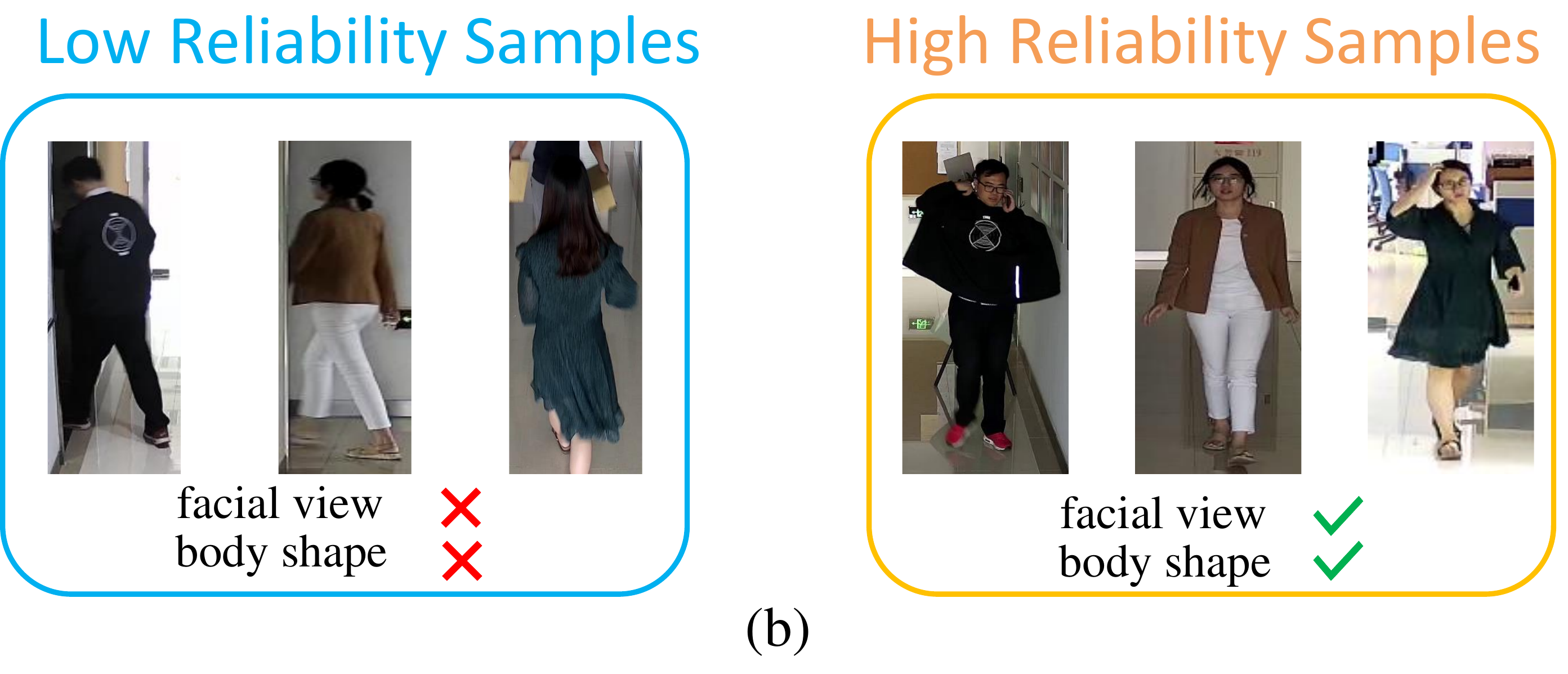}
        \label{fig:reliability examples}
    \end{subfigure}
    \caption{(a) Illustrative examples of intermediary availability. When matching with clothes-relevant features, we consider the availability as the accessibility of same-clothes samples. When matching with clothes-irrelevant features, we consider the availability as the accessibility of same-identity samples. (b) Illustrative examples of intermediary reliability. High reliability samples usually have clear clothes-irrelevant cues (facial view and body shape), while low reliability samples typically lack integrity in face and body shape.}
    \label{fig:availability and reliability example}
\end{figure}
 
 Built upon this inspiration, we employ clothes-relevant features to establish a novel intermediary matching process. We introduce the intermediary sample, which refers to an extra data sample that serves as a median between the matching of query and target. In this way, instead of directly matching samples with insufficient clothes-irrelevant information to the target, we can first match them to intermediary samples that are more informative, and then match these intermediaries to target. Specifically, as depicted in Fig.~\ref{fig: idea illustration}, for query ($A$) lacking adequate identity information, we can leverage clothes-relevant features to match the query with intermediaries that have richer clothes-irrelevant identity characteristics (\textit{e.g.}, face). For query ($B$) exhibiting large intra-class variation with target ($B$), the clothes-relevant features can be utilized to match the query with intermediaries that have better-aligned clothes-irrelevant clues (\textit{e.g.}, body shape) to the target. This intermediary matching approach streamlines the matching between the query and target individuals, utilizing intermediaries with highly informative clothes-irrelevant characteristics. Notably, given that images of pedestrians wearing the same clothes are easily accessible from nearby frames or multiple camera views in short-term video surveillance, the fundamental feasibility of the intermediary matching process is guaranteed. 
 
 Nevertheless, the intermediary matching process would be less effective if the desired intermediaries (\textit{e.g.}, same-clothes samples) are unavailable, or if the retrieved intermediaries contain unreliable clothes-irrelevant identity information. Under these conditions, the re-id performance may still be impaired. Therefore, it becomes imperative to assess the \textit{feasibility} of intermediary matching, which refers to how well the matching process could be effectively performed with the fetched intermediaries. To judge the feasibility of intermediary matching process, we propose to evaluate the quality of intermediaries, by considering both their \textit{availability} and the \textit{reliability of their clothes-irrelevant identity information}: (1) \textbf{Availability} refers to the accessibility of the desired intermediary samples. Specifically, when matching with clothes-relevant features, availability refers to the accessibility of same-clothes intermediaries. For example, as shown on the left part of Fig.~\ref{fig:availability and reliability example}(a), if same-clothes samples do not exist in the data source of intermediaries, the availability of intermediaries is low. On the other hand, when matching with clothes-irrelevant features, availability refers to the accessbility of same-identity samples. As shown on the right part of Fig.~\ref{fig:availability and reliability example}(a), if same-identity samples do not exist in the data source, the availability of intermediaries is low as well. (2) \textbf{Reliability} refers to the integrity of clothes-irrelevant identity cues in the intermediary samples. For example, as shown in Fig.~\ref{fig:availability and reliability example}(b), intermediaries with clear facial views and full body views possess high reliability, whereas intermediaries with obscured face or occluded body parts have low reliability.

To explore richer identity clues through intermediary matching while also addressing the varying quality of intermediaries, in this paper, we propose a Feasibility-Aware Intermediary Matching (FAIM) framework for clothes-changing re-id. FAIM contains an Intermediary Matching (IM) module that conducts multiple intermediary-assisted matching routes utilizing both clothes-relevant and clothes-irrelevant features. By jointly exploiting these matching routes, IM module can tackle all scenarios where clothes-irrelevant features of query or target samples lack sufficient identity information. Furthermore, an Intermediary-Based Feasibility Weighting (IBFW) module is designed to assign feasibility weights to different intermediary matching routes. These weights are determined based on the availability of intermediaries and the reliability of clothes-irrelevant identity information carried by them. On one hand, the availability of intermediaries can be assessed by feature similarities, \textit{e.g.}, intermediaries that have lower clothes-relevant feature similarity to query are more likely to be suboptimal samples when same-clothes samples are deficient, indicating low availability. On the other hand, we train an Identity Information Reliability (IIR) module to learn the reliability of clothes-irrelevant identity information and use it to predict the reliability score of intermediaries when doing IM. 
Extensive experiments show that our framework outperforms other methods on several clothes-changing re-id benchmarks, including LTCC~\cite{qian2020long}, PRCC~\cite{yang2019person} and DeepChange~\cite{xu2023deepchange}, demonstrating the superiority of our framework.

\section{Related Works}

\subsection{Clothes-changing person re-id}
In the original person re-identification task, it is assumed that people do not change their clothes across all time periods. Based on this traditional setting, a series of more challenging tasks are derived. Visible-infrared re-id aims at retrieving the same identity across RGB and infrared modalities. ~\cite{li2022visible} proposed a modality-specific memory network to learn a more unified feature for cross-modality retrieval. Occluded re-id attempts to tackle situations where people are only partly visible. ~\cite{li2021diverse} treated this problem by designing a part-aware transformer to learn representative part features for identities. Moreover,~\cite{wu2023learning} focuses on cross-resolution re-id, which aims at matching person images with varying resolutions. This work addresses this task by learning a resolution-adaptive representation. ~\cite{wu2022pseudo} studies the problem of unsupervised re-id by raising a self-similarity learning approach that learns discriminative features by pseudo pairs.

Among these challenging settings, clothes-changing person re-id aims at retrieving same-identity person when they expose to clothes changes. Current clothes-changing person re-id methods~\cite{gu2022clothes,hong2021fine,huang2019beyond,jin2022cloth,shu2021semantic, han2023clothing} focus on extracting clothes-irrelevant identity features. A line of works resorts to multi-modality data, such as contour sketch~\cite{yang2019person, chen2021deep}, human silhouette~\cite{hong2021fine, cui2023dcr, zhao2018understanding}, radio signal~\cite{fan2020learning}, body keypoints~\cite{qian2020long} and 3D shape~\cite{chen2021learning, liu2023learning}. For instance, the work~\cite{qian2020long} develops a clothes-elimination shape-distillation framework to extract clothes-irrelevant representations under the guidance of body keypoint embeddings, and the work~\cite{cui2023dcr} disentangles clothes-irrelevant cues from clothes-relevant cues by letting the model to reconstruct the silhouettes of clothes-irrelevant and clothes-relevant parts separately. Another line of work leverages facial information~\cite{wang2021face, zhao2018towards} to aid re-id in clothes-changing scenarios. Thanks to the rapid development in face detection and recognition techniques~\cite{zhao20183d, zhao2020towards, deng2020retinaface, zhang2016joint}, the works~\cite{yu2020cocas, wu2022identity} detect and extract facial representations to serve as a strongly reliable identity information for retrieval. Another type of method extract clothes-irrelevant features using single-modality image data. For example, the work~\cite{gu2022clothes} relies on adversarial learning to mine clothes-irrelevant information, the work~\cite{han2023clothing} conducts a clothes-change ID-unchange feature augmentation to boost the decomposition of clothes-independent identity information, and the work~\cite{liu2023clothes} proposes an association-forgetting method to facilitate the learning of identity-relevant features through association learning, while precluding the impact of identity-irrelevant features through forgetting learning. All the above works assume that the \textit{clothes-irrelevant} feature alone is sufficient to represent pedestrians. However, the large variations in human pose and camera view can cause clothes-irrelevant features to lack identity clues and exhibit large intra-class variations. In this work, we propose an intermediary-assisted matching strategy, which additionally utilizes \textit{clothes-relevant} characteristic to reduce identification difficulty.

 \subsection{Re-ranking in re-id}
 Re-ranking is a useful tool to improve performance on retrieval tasks~\cite{qin2011hello, chum2007total,li2012common,tan2021instance,ye2016person} by leveraging extra samples from gallery. Recently, neighbor-similarity re-ranking works, such as  \textit{k}-reciprocal encoding~\cite{bai2019re,zhong2017re} and graph neural networks~\cite{zhang2020understanding}, have achieved satisfactory performance on re-id without clothes changing. However, these re-ranking methods perform matching within a single feature space. In clothes-changing re-id, the current strategies may fail, as intra-identity features could be far from each other in a single feature space. Differently, our IM \textit{indirectly} matches images across both \textit{clothes-relevant} and \textit{clothes-irrelevant} feature spaces, enabling better matching in clothes-changing re-id.

\subsection{Reliability modeling in re-id}
In re-id task, a group of works~\cite{sun2021part, zheng2021exploiting, yu2019robust, dou2022reliability, jin2020uncertainty} model the reliability of data samples to mitigate the impact of outliers or noisy labels. In weakly-supervised setting,~\cite{yang2022uncertainty} proposes to utilize reliability-aware methodology to estimate the uncertainty of self-generated pseudo labels. Recent works~\cite{yu2019robust, dou2022reliability} predict the sample reliability by mapping each sample to a Gaussian distribution in latent feature space, and learning the variance which represents sample reliability. These works only model the reliability for same-clothes scenarios, and the predicted reliability is only used for training. Differently, our method builds the reliability for clothes-changing scenarios, while managing to apply the reliability score for the testing procedure.

\begin{figure*}[h!t]
\centering
    \includegraphics[width=0.85\linewidth]{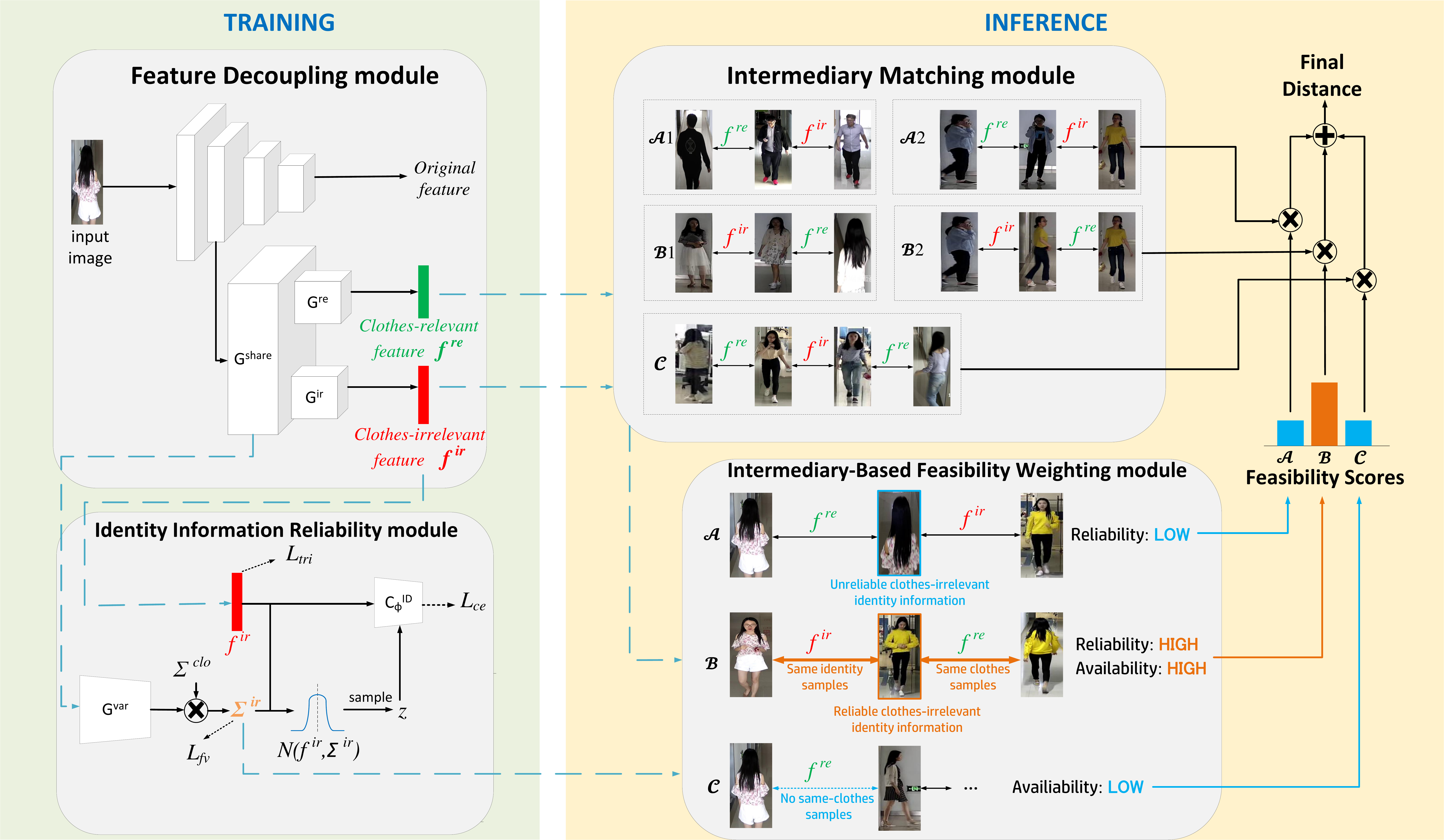}
    \caption{Overview of our FAIM framework. (1) A Feature Decoupling module is utilized to extract clothes-relevant feature $\boldsymbol{f}^{re}$ and clothes-irrelevant feature $\boldsymbol{f}^{ir}$. (2) An Identity Information Reliability module is designed to predict reliability score of identity information in $\boldsymbol{f}^{ir}$. (3) An Intermediary Matching module is proposed, which conducts three matching routes $\mathcal{A},\mathcal{B},\mathcal{C}$ to comprehensively address situations where direct matching solely based on clothes-irrelevant features is ineffective. (4) An Intermediary-Based Feasibility Weighting module is utilized to assign feasibility weights to routes $\mathcal{A}\sim \mathcal{C}$ respectively, according to the availability and reliability of intermediaries.}
    \label{fig: method}
\end{figure*}
\vspace{-3mm}

\section{Method}

The framework of our method is presented in Fig.~\ref{fig: method}. In the training stage, a Feature Decoupling (FD) module is trained to output three features: the original image feature $\boldsymbol f^{o}$, clothes-irrelevant feature $\boldsymbol f^{ir}$ and clothes-relevant feature $\boldsymbol f^{re}$. An Identity Information Reliability (IIR) module is jointly trained with the FD module to predict the reliability of clothes-irrelevant identity information, $r^{id}$. In the inference stage, the Intermediary Matching (IM) module utilizes $\{\boldsymbol{f}^{o}, \boldsymbol{f}^{ir}, \boldsymbol{f}^{re}\}$ to conduct intermediary-assisted matching through multiple routes and calculate the intermediary-based distance for each route. After that, an Intermediary-Based Feasibility Weighting (IBFW) module is deployed to dynamically re-weight different routes, and obtain the final matching results. In the next subsections, we will elaborate each module in detail.

\subsection{Feature Decoupling Module}
\label{sec: feature decoupling module}
As shown in Fig.~\ref{fig: method}, in the feature decoupling module, we deploy a dual-branch structure to derive the clothes-irrelevant feature $\boldsymbol{f}^{ir}$ and clothes-relevant feature $\boldsymbol f^{re}$. Specifically, given an input image $\boldsymbol{x}$, we first fed it to a $l$-layer deep network $F_{\theta}\left(\cdot\right)$ to get an original image feature $\boldsymbol{f}^{o}=F_{\theta}\left(\boldsymbol{x}\right)$. Meanwhile, the output feature map of the $(l-2)$-th layer, $\boldsymbol{f}^{(l-2)}$, is passed through a shared convolutional block $G^{share}$, and then passed through two separate convolutional blocks $G^{ir}$, $G^{re}$ to derive $\boldsymbol{f}^{ir}$ and $\boldsymbol{f}^{re}$.

To enable $\boldsymbol{f}^{ir}$ to extract clothes-irrelevant identity information, we utilize the \textit{identity category} $y^{ID}$ to supervise its learning. In addition, to enable $\boldsymbol{f}^{re}$ to extract clothes-relevant identity information, we use \textit{clothes category} $y^{C}$, where each identity category is further divided into multiple fine-grained categories of clothes, to supervise its learning. Specifically, following \cite{hong2021fine, huang2021clothing}, we optimize  $\boldsymbol{f}^{ir}$ using a combination of identity classification loss (cross-entropy loss) $\ell_{ce}$ and triplet loss $\ell_{tri}$, which are both conditioned on \textit{identity category} $y^{ID}$:
\begin{equation}
    \mathcal{L}_{ir}=\ell_{ce}\left(C_{\phi}^{ID}\left(\boldsymbol{f}^{ir}\right), y^{ID}\right)+\ell_{tri}\left(\boldsymbol{f}^{ir}\right | y^{ID}),
    \label{eq: clothes-irrelevant feature loss}
\end{equation}
where $C_{\phi}^{ID}\left(\cdot\right)$ denotes the identity classifier with parameter $\phi$. Similarly, we use clothes classification loss and triplet loss conditioned on \textit{clothes category} $y^{C}$ to optimize $\boldsymbol{f}^{re}$:
\begin{equation}
    \mathcal{L}_{re}=\ell_{ce}\left(C_{\varphi}^{C}\left(\boldsymbol{f}^{re}\right), y^{C}\right)+\ell_{tri}\left(\boldsymbol{f}^{re}\right | y^{C}),
    \label{eq: clothes-relevant feature loss}
\end{equation}
where $C_{\varphi}^{C}\left(\cdot\right)$ denotes the clothes classifier with parameter $\varphi$. Notably, in existing clothes-changing re-ID datasets, since clothes labels are fine-grained identity labels, samples from different identities have different clothes labels even if they actually wear same clothes. Therefore, to more effectively mine fine-grained clothes information in $\ell_{tri}\left(\boldsymbol{f}^{re}\right | y^{C})$, we only select negative samples within the same identity category as the anchor sample, to guarantee that their clothes are not same. Specifically, the formulation of $\ell_{tri}\left(\boldsymbol{f}^{re}\right | y^{C})$ is as follows:
\begin{align}
    \ell_{tri}\left(\boldsymbol{f}^{re} | y^{C}\right) = \frac{1}{B}&
    \sum_{a} \left[d(\boldsymbol{f}^{re}_a, \boldsymbol{f}^{re}_p) - d(\boldsymbol{f}^{re}_a, \boldsymbol{f}^{re}_n) + m\right]_{+} \nonumber \\
    p = \arg\max_i & \,d(\boldsymbol{f}^{re}_a, \boldsymbol{f}^{re}_i), s.t. \,y^{C}_i = y^{C}_a. \nonumber \\
    n = \arg\min_i & \,d(\boldsymbol{f}^{re}_a, \boldsymbol{f}^{re}_i), s.t. \,y^{C}_i \neq y^{C}_a,\quad y^{ID}_i = y^{ID}_a.
    \label{eq: clothes triplet loss}
\end{align}
where $\{a, p, n\}$ denotes a triplet from a mini-batch with batch size $B$: $a$ is anchor sample, $p$ is the positive sample with the largest distance from $a$, and $n$ is the negative sample with smallest distance from $a$. $d$ is cosine distance metric. $\left[\cdot\right]_+$ denotes $\max(\cdot, 0)$. $y^{C}$ denotes the clothes label and $y^{ID}$ denotes the identity label.

\subsection{Identity Information Reliability Module}
\label{sec: identity information reliability block}
Prior works~\cite{yu2019robust, dou2022reliability} model the reliability of data by mapping the data sample to a Gaussian distribution in the latent space, where a lower variance refers to higher sample reliability. Different from these approaches that directly learn the variance, we construct a clothes-changing variance to model the reliability of \textit{\textbf{clothes-irrelevant}} identity information. As shown in Fig.~\ref{fig: method}, following ~\cite{yu2019robust}, we map the feature $\boldsymbol{f}^{ir}$ to the Gaussian distribution $\mathcal{N}(\boldsymbol{f}^{ir}, \boldsymbol{\Sigma}^{ir})$ by drawing $N$ random samples $\{\boldsymbol{z}_j\}_{j=1}^N$ where $\boldsymbol{z}_j=\boldsymbol{f}^{ir} + \boldsymbol{\epsilon}_j\boldsymbol{\Sigma}^{ir}, \boldsymbol{\epsilon}_j\in\mathcal{N}(0, \boldsymbol{I})$. Based on semantic transformations~\cite{wang2019implicit, han2023clothing}, we obtain $\boldsymbol{\Sigma}^{ir}=\rho\cdot\boldsymbol{\Sigma}^{clo}$ from clothes-changing semantic directions, where  $\rho=G^{var}(G^{share}(\boldsymbol{f}^{(l-2)}))$\footnote{$G^{var}$ is a convolutional block.} is a learnable scalar that controls the scale of $\boldsymbol{\Sigma}^{ir}$, and $\boldsymbol{\Sigma}^{clo}$ is an instance-wise clothes-changing variance that controls the direction of $\boldsymbol{\Sigma}^{ir}$. In particular, the clothes-changing variance for sample $i$ is derived as follows:
\begin{equation}
    \boldsymbol{\Sigma}^{clo}_i = norm(\mathbb{E}_{i\neq j}\left[(\boldsymbol{\mu}_{y_i}^{c_i} - \boldsymbol{f}_{y_i}^{c_j})^2\right]),
\end{equation}
where $\boldsymbol{\mu}_{y_i}^{c_i}$ is the feature center with identity label $y_i$ and clothes label $c_i$, $\boldsymbol{f}_{y_i}^{c_j}$ is the feature sample with identity label $y_i$ and clothes label $c_j$ $(c_j\neq c_i)$, and $norm$ stands for $l_2$-normalization. To learn the reliability, we jointly optimize $\boldsymbol{z}_j$ and $\boldsymbol{\Sigma}^{ir}$ by reformulating $\ell_{ce}$ in Eq.~\ref{eq: clothes-irrelevant feature loss} as:
\begin{equation}
    \mathcal{L}_{cls}=\ell_{ce}(C_{\phi}^{ID}(\boldsymbol{f}^{ir}), y^{ID}) + \frac{1}{N}\sum^{N}_{j=1}\ell_{ce}(C_{\phi}^{ID}(\boldsymbol{z}_j), y^{ID}),
    \label{eq: new clothes-irrelevant feature classification loss}
\end{equation}
In this process, to maintain low classification loss, $\boldsymbol{f}^{ir}$ with high reliability will obtain a low $\boldsymbol{\Sigma}^{ir}$ to constrain the sampled features close to itself. However, as lower variance will lead to higher consistency between the sampled features and reduce the classification loss, a shortcut is to decrease $\boldsymbol\Sigma^{ir}$ of all samples to zero. To avert this shortcut, we add a feature variance loss~\cite{yu2019robust} to maintain the average variance across all training samples at a certain level:
\begin{equation}
    \mathcal{L}_{fv}=\max(0, \lambda_{fv} - \rho),
    \label{eq: feature variance loss}
\end{equation}
 where $\lambda_{fv}$ is the margin to bound the average variance level. By employing $L_{fv}$, the model will have to find a balance between reducing the classification loss and maintaining a total variance level. This will lead the model to reduce the variance of high-reliability samples and maintain the variance of low-reliability samples. Thus we can ensure that samples with higher reliability exhibit low feature variance. In this way, we can predict the reliability score of clothes-irrelevant identity information as $r^{id} = 1.0 - \rho$.

At last, we reformulate $\mathcal{L}_{ir}$ in Eq.~\ref{eq: clothes-irrelevant feature loss} as $\mathcal{L}_{ir}=\mathcal{L}_{cls}+\mathcal{L}_{fv}+\ell_{tri}\left(\boldsymbol{f}^{ir}\right | y^{ID})$, and the total loss function of the Feature Decoupling module is formed as follows:
\begin{equation}
    \label{eq: total loss} \mathcal{L}=\mathcal{L}_o + \alpha_{ir}\mathcal{L}_{ir} + \alpha_{re}\mathcal{L}_{re},
\end{equation}
where $\mathcal{L}_o$ consists of the identity classification loss and triplet loss with the original feature $\boldsymbol{f}^o$, and $\alpha_{ir}$ and $\alpha_{re}$ are the hyper-parameters to balance different loss functions.

\subsection{Intermediary Matching Module}
\label{sec: intermediary matching module}
When clothes-irrelevant characteristic lacks sufficient identity clues, direct matching between query and target samples would become intractable. To this end, we propose an IM module that performs an intermediary-assisted matching process by conducting multiple matching routes jointly using clothes-irrelevant and clothes-relevant features.

\

\textbf{Intermediary-assisted Matching Routes.} As discussed in Sec.~\ref{sec: introduction}, clothes-irrelevant features could suffer from inadequate identity information and large intra-class variations. \textbf{\textit{To address the inadequate identity information challenge}}, we deal with two possible cases: (1) Only the query (or target) lacks clothes-irrelevant identity information, as shown in case $\mathcal{A}1$ (or $\mathcal{B}1$) in Fig.~\ref{fig: method}. In this case, we can use \textit{clothes-relevant features} to match the query (or target) to informative intermediate samples, and then use \textit{clothes-irrelevant features} to match the intermediate samples to the target (or query). (2) Both query and target lack clothes-irrelevant identity information, as shown in case $\mathcal{C}$ in Fig.~\ref{fig: method}. In this case, we can first match query and target to informative intermediaries respectively through \textit{clothes-relevant features}, and then match the intermediaries through \textit{clothes-irrelevant features}. \textbf{\textit{To address the large intra-class variations challenge}}, we can utilize \textit{clothes-relevant features} to match the query (or target) to intermediaries with aligned pose and view information to target (or query), and then utilize \textit{clothes-irrelevant features} to match these intermediaries to the target (or query), as shown in case $\mathcal{A}2$ (or $\mathcal{B}2$) in Fig.~\ref{fig: method}. In summary, we can construct three intermediary-assisted matching routes $\left\{\mathcal{A},\mathcal{B},\mathcal{C}\right\}$ to more effectively identify clothes-changing  pedestrians: 
\begin{align}
    &\mathcal{A}: \ q \stackrel{f^{re}}{\longleftrightarrow} i \stackrel{f^{ir}}{\longleftrightarrow} t, \nonumber \\
    &\mathcal{B}: \ q \stackrel{f^{ir}}{\longleftrightarrow} i \stackrel{f^{re}}{\longleftrightarrow} t, \nonumber \\
    &\mathcal{C}: \ q \stackrel{f^{re}}{\longleftrightarrow} i_1 \stackrel{f^{ir}}{\longleftrightarrow} i_2 \stackrel{f^{re}}{\longleftrightarrow} t.
  \label{eq: intermediary matching routes}
\end{align}
where $q$, $t$ and $i$ denote query, target and intermediary samples respectively. $q \stackrel{f^{re}}{\longleftrightarrow} i$ denotes the \textit{intermediary matching path} that matches query $q$ to the intermediate $i$ using $\boldsymbol{f}^{re}$, analogously for other matching paths.

\

\textbf{Intermediary-based Distance.} To measure the distance of intermediary-assisted matching routes, we devise an \textit{intermediary-based distance} based on two mainstream re-ranking methods, namely \textit{k}-reciprocal encoding~\cite{zhong2017re} and Graph Neural Network (GNN)-based~\cite{zhang2020understanding}.

\

\noindent{\textbf{\textit{k-Reciprocal Encoding IM.}}}
Following \textit{k}-reciprocal encoding re-ranking~\cite{zhong2017re}, we design a Jaccard metric to measure the intermediary-based distance. We take route $\mathcal{A}$ in Eq.~\ref{eq: intermediary matching routes} as an example. In particular, we obtain the \textit{k}-reciprocal neighbors $R^{re}\left(q,k\right)$ of query $q$ based on clothes-relevant feature $\boldsymbol{f}^{re}$, and the \textit{k}-reciprocal neighbors $R^{ir}\left(t,k\right)$ of target $t$ based on clothes-irrelevant feature $\boldsymbol{f}^{ir}$. Then, the intermediary matching distance $d_\mathcal{A}\left(q, t\right)$ is computed as:
\begin{equation}
    d_{\mathcal{A}}\left(q, t\right)=1-\frac{\left|R^{re}\left(q,k\right) \cap R^{ir}\left(t,k\right)\right|}{\left|R^{re}\left(q,k\right) \cup R^{ir}\left(t,k\right)\right|}.
    \label{eq:6}
\end{equation}
a smaller value of $d_{\mathcal{A}}\left(q, t\right)$ indicates a higher overlap proportion of intermediaries between $q$ and $t$, which suggests a closer distance along matching route $\mathcal{A}$. Analogously, the distance along routes $\mathcal{B}$ and $\mathcal{C}$ can be computed as:
\begin{gather}
    d_{\mathcal{B}}\left(q, t\right)=1-\frac{\left|R^{ir}\left(q,k\right) \cap R^{re}\left(t,k\right)\right|}{\left|R^{ir}\left(q,k\right) \cup R^{re}\left(t,k\right)\right|}, \nonumber \\
    d_{\mathcal{C}}\left(q,t\right)=1-\frac{\left|R^{\mathcal{A}}\left(q,k\right) \cap R^{re}\left(t,k\right)\right|}{\left|R^{\mathcal{A}}\left(q,k\right) \cup R^{re}\left(t,k\right)\right|},
    \label{eq:7}
\end{gather}
where $R^{\mathcal{A}}\left(q,k\right)$ denotes the \textit{k}-reciprocal neighbors of $q$, measured by $d_{\mathcal{A}}$.

\

\noindent\textbf{\textit{GNN-Based IM}}.
Following GNN-based re-ranking~\cite{zhang2020understanding}, we construct a neighbor encoding feature to perform intermediary matching process.   Take the matching path $\mathcal{A}$ in Eq.~\ref{eq: intermediary matching routes} as an example. First, based on $\boldsymbol{f}^{re}$, we construct a distance matrix $\boldsymbol{D}^{re} \in \mathbb{R}^{n\times n}$ of query and gallery set with $n$ samples in total. Next, we derive a $n$-dimensional  neighbor vector $\boldsymbol{g}^{re}_i=\left[g^{re}_{i,1},g^{re}_{i,2}, \dots, g^{re}_{i,n} \right]$ for each sample $i$, where $g^{re}_{i,j}$ is computed as follows:
\begin{equation}
    g^{re}_{i,j}=\left\{
    \begin{aligned}
        1,  & \qquad  
            \text{if} \ j \in \mathcal{N}^{re}\left(i,k\right) \land i \in \mathcal{N}^{re}\left(j,k\right), \\
        0,  & \qquad
            \text{if} \ j \notin \mathcal{N}^{re}\left(i,k\right) \land i \notin \mathcal{N}^{re}\left(j,k\right), \\
        0.5, & \qquad \text{otherwise}
    \end{aligned}
    \right.
    \label{eq:10}
\end{equation}
where $\mathcal{N}^{re}\left(i,k\right)$ denotes the \textit{k}-nearest neighbors of sample $i$, obtained based on $\boldsymbol{D}^{re}$. Then, following~\cite{gilmer2017neural}, the GNN-based clothes-relevant neighbor encoding feature $\boldsymbol{h}^{re}_i$ is obtained from $\boldsymbol{g}^{re}_i$:
\begin{equation}
    \boldsymbol{h}^{re}_i = \boldsymbol{g}^{re}_i + \sum\nolimits_j \left({e^{re}_{ij}}\right)^2\cdot \boldsymbol{g}^{re}_{j}.
    \label{eq:11}
\end{equation}
Here, the edge weight $e^{re}_{ij}$ is the cosine similarity between $\boldsymbol{f}^{re}_i$ and $\boldsymbol{f}^{re}_j$, and the square operation is used to further enhance edges with high weights. Likewise, we can derive GNN-based clothes-irrelevant neighbor encoding feature $\boldsymbol{h}^{ir}_i$. Finally, the distance of route $\mathcal{A}$ is obtained by the cosine distance of $\boldsymbol{h}^{re}$ and $\boldsymbol{h}^{ir}$:
\begin{equation}
    d_{\mathcal{A}}\left(q,t\right) = -\cos\left(\boldsymbol{h}^{re}_q, \boldsymbol{h}^{ir}_{t}\right).
    \label{eq:12}
\end{equation}
and the distance of routes $d_{\mathcal{B}}$ and $d_{\mathcal{C}}$ can be computed as:
\begin{align}
    & d_{\mathcal{B}}\left(q,t\right) = -\cos\left(\boldsymbol{h}^{ir}_q, \boldsymbol{h}^{re}_{t}\right) \nonumber \\
    & d_{\mathcal{C}}\left(q,t\right) = -\cos\left(\boldsymbol{h}^{A}_q, \boldsymbol{h}^{re}_{t}\right).
    \label{eq:13}
\end{align}
Note that $d_{\mathcal{C}}$ is derived by first acquiring $\boldsymbol{h}^{A}_q$, where $\boldsymbol{h}^{A}_q$ denotes the neighbor encoding feature of matching path $\mathcal{A}$ based on $d_{\mathcal{A}}$.

\

\begin{table*}[tbp]
\centering
\caption{Comparison with state-of-the-arts on LTCC~\cite{qian2020long}, PRCC~\cite{yang2019person} and DeepChange~\cite{xu2023deepchange} benchmarks. `general', `SC' and `CC' denotes the three evaluation protocols illustrated in Sec.~\ref{sec: datasets and evaluation protocols}. `k-r' and `GNN' denotes results of employing IM with \textit{k-reciprocal}~\cite{zhong2017re} and \textit{GNN-based}~\cite{zhang2020understanding} methods, respectively. The best performance under each setting is boldfaced, while the second-best performance under each setting is underlined. `-' denotes not reported in original paper.}
\scalebox{1.1}{
\begin{tabular}{l|c|c|cc|cc|cc|cc|cc}
\hline
    \multirow{3}*{method} & \multirow{3}*{modality} & \multirow{3}*{reference} & \multicolumn{4}{c|}{LTCC} & 
        \multicolumn{4}{c|}{PRCC} & \multicolumn{2}{c}{DeepChange} \\
    \cline{4-13}
    &&& \multicolumn{2}{c|}{general} & \multicolumn{2}{c|}{CC} &
    \multicolumn{2}{c|}{SC} & \multicolumn{2}{c|}{CC} & \multicolumn{2}{c}{general} \\
    \cline{4-13}
    &&& top-1 & mAP & top-1 & mAP & top-1 & mAP & top-1 & mAP & top-1 & mAP \\
\hline
    IANet \cite{hou2019interaction} & RGB & CVPR'19 & 63.7 & 31.0 & 25.0 & 12.6 & 99.4 & 98.3 & 46.3 & 45.9 & - & - \\
    OSNet \cite{zhou2019omni} & RGB & ICCV'19 & 66.1 & 31.1 & 23.4 & 10.3 & - & - & - & - & - & - \\
\hline
    SPT+ASE \cite{yang2019person} & RGB+sketch & TPAMI'19 & - & - & - & - & 64.2 & - & 34.4 & - & - & - \\
    GI-ReID \cite{jin2022cloth} & RGB+sil. & CVPR'22 & 63.2 & 29.4 & 23.7 & 10.4 & 80.0 & - & 33.3 & - & - & - \\
    CESD \cite{qian2020long} & RGB+pose & ACCV'20 & 71.4 & 34.3 & 26.2 & 12.4 & - & - & - & - & - & - \\
    FSAM \cite{hong2021fine} & RGB+sil. & CVPR'21 & 73.2 & 35.4 & 38.5 & 16.2 & 98.8 & - & 54.5 & - & - & - \\
    MAC-DIM \cite{chen2021deep} & RGB+contour & TMM'21 & 70.9 & 34.0 & 29.9 & 13.0 & 95.2 & - & 48.8 & - & - & - \\
    CAL \cite{gu2022clothes} & RGB & CVPR'22 & 74.2 & 40.8 & 40.1 & 18.0 & 100 & 99.8 & 55.2 & 55.8 & 54.0 & 19.0 \\
    AIM \cite{yang2023good} & RGB & CVPR'23 & 76.3 & 41.1 & 40.6 & 19.1 & 100 & 99.9 & 57.9 & 58.3 & - & - \\
    CCFA \cite{han2023clothing} & RGB & CVPR'23 & 75.8 & 42.5 & 45.3 & 22.1 & 99.6 & 98.7 & \underline{61.2} & 58.4 & - & - \\
    3DInvarReID \cite{liu2023learning} & RGB+pose & ICCV'23 & - & - & 40.9 & 18.9 & - & - & 56.5 & 57.2 & - & - \\
    DCR-ReID \cite{cui2023dcr} & RGB+sil. & TCSVT'23 & 76.1 & 42.3 & 41.1 & 20.4 & 100 & 99.7 & 57.2 & 57.4 & - & - \\
    AFL \cite{liu2023clothes} & RGB & TMM'23 & 74.4 & 39.1 & 42.1 & 18.4 & 100 & 99.7 & 57.4 & 56.5 & - & - \\
    SCNet \cite{guo2023semantic} & RGB+sil. & ACM MM'23 & 76.3 & 43.6 & 47.5 & 25.5 & 100 & 97.8 & \textbf{61.3} & 59.9 & 53.5 & 18.7 \\
    MCSC-CAL \cite{huang2024meta} & RGB & TIP'24 & 73.9 & 40.2 & 42.2 & 19.4 & 99.8 & 99.8 & 57.8 & 57.3 & 56.9 & 21.5 \\
\hline
    Baseline & RGB & This paper & 73.4 & 36.8 & 36.0 & 16.0 & 100 & 99.8 & 53.4 & 53.8 & 53.8 & 18.1 \\
    
    \textbf{FAIM(k-r)} & RGB & This paper & \textbf{79.5} & \textbf{53.4} & \textbf{48.2} & \textbf{27.5} & \textbf{100} & \textbf{100} & 60.9 & \underline{62.0} & \textbf{61.5} & \textbf{28.7} \\
    \textbf{FAIM(GNN)} & RGB & This paper & \underline{78.1} & 
    \underline{48.6} & \underline{46.2} & \underline{26.0} & \underline{100} & \underline{100} & 59.8 & \textbf{62.5} & \underline{58.9} & \underline{26.6} \\
  \hline
\end{tabular}}
\label{tab: comparison with sota}
\end{table*}

\subsection{Intermediary-Based Feasibility Weighting Module}
\label{sec: intermediary-based feasibility weighting}
Though IM module can help bridge the gap between query and gallery samples, it is worth noting that both the availability and clothes-irrelevant identity information reliability can affect the feasibility of intermediary-assisted matching routes. To this end, as shown in Fig.~\ref{fig: method}, we design an Intermediary-Based Feasibility Weighting approach to assign feasibility weights for each matching route $\left\{\mathcal{A},\mathcal{B},\mathcal{C}\right\}$, by comprehensively measuring the availability and reliability of the intermediaries found via each route.

Take route $\mathcal{A}$ as an example. Firstly, we consider two aspects  to evaluate the availability of intermediaries: (1) When matching intermediaries from $q$ with $\boldsymbol{f}^{re}$, it is crucial to consider whether intermediaries with same clothes as $q$ is available. Generally, given that intermediaries represent the top nearest neighbors of $q$ in clothes-relevant feature space, a higher similarity in clothes-relevant features between query and intermediaries indicates a high availability of intra-identity same-clothes intermediaries. (2) Likewise, a higher similarity of $\boldsymbol{f}^{ir}$ indicate high availability of intra-identity clothes-changing intermediaries when matching target $t$ with intermediaries. Secondly, since clothes-irrelevant features are employed to match $t$ with intermediaries, it is crucial to consider the reliability of clothes-irrelevant identity information. Consequently, given query $q$, target $t$ and the intermediary set $\mathcal{I}$, we can compute the feasibility score of route $\mathcal{A}$ as follows:
\begin{equation}
    s_{\mathcal{A}}(q,t) = \frac{1}{|\mathcal{I}|}\sum_{i\in\mathcal{I}}s^{re}(q,i)\cdot s^{ir}(i,t)\cdot r^{id}(i)
\end{equation}
where $s^{re}$ and $s^{ir}$ denotes the cosine similarity (scaled to $\left[0,1\right]$) of $\boldsymbol{f}^{re}$ and $\boldsymbol{f}^{ir}$, respectively. $r^{id}$ denotes the reliability score computed by the IIR module. Similarly, we can compute $s_\mathcal{B}$ and $s_\mathcal{C}$:
\begin{equation}
\begin{split}
    s_{\mathcal{B}}(q,t) &= \frac{1}{|\mathcal{I}|}\sum_{i\in\mathcal{I}}s^{ir}(q,i)\cdot s^{re}(i,t)\cdot r^{id}(i), \\
    s_{\mathcal{C}}(q,t) &= \frac{1}{|\mathcal{I}|}\sum_{i\in\mathcal{I}}s^{\mathcal{A}}(q,i)\cdot s^{re}(i,t)\cdot r^{id}(i).
\end{split}
\end{equation}
Eventually, the final distance function $d^*$ after feasibility re-weighting can be formulated as:
\begin{equation}
\begin{split}
    &d^*\left(q,t\right)=d_{I}\left(q,t\right) + \lambda_{o}\left[d\left(q,t\right) + d_{o}\left(q,t\right)\right],  \\
    &\text{where,} \quad d_{I}(q,t)=s_{\mathcal{A}}d_{\mathcal{A}}+s_{\mathcal{B}}d_{\mathcal{B}}+s_{\mathcal{C}}d_{\mathcal{C}}, \\
    &\lambda_o = 1 - \frac{s_{\mathcal{A}} + s_{\mathcal{B}} + s_{\mathcal{C}}}{3},
    \label{eq: final distance}
\end{split}
\end{equation}
where $d$ is the original cosine distance, and $d_o$ is the original Jaccard distance. Both $d$ and $d_o$ are computed based on original feature $\boldsymbol{f}^o$.

\section{Experiments}
In this section, we evaluate FAIM on several benchmarks and conduct ablation studies to validate the effectiveness of major components. More implementation details and experimental results are shown in supplementary material.

\subsection{Datasets and Evaluation Protocols}
\label{sec: datasets and evaluation protocols}
\textbf{Datasets.} \
We mainly evaluate our framework on three widely-used clothes-changing re-id benchmarks, \textit{i.e.} LTCC~\cite{qian2020long}, PRCC~\cite{yang2019person} and DeepChange~\cite{xu2023deepchange}. LTCC~\cite{qian2020long} is a long-term person re-id dataset that covers images of various outfits for each individual. The dataset contains 17,119 images of 152 IDs in total, where 14,783 images of 91 IDs have more than one outfits (with 416 outfits in total). Images are captured across long periods of time (up to 2 months), and with up to 12 different camera views.
PRCC~\cite{yang2019person} is a clothes-changing person re-id dataset with 33,698 images of 221 IDs in total. Images of each identity consist of two different outfits taken from three different camera views. The samples taken from camera A and B share the same clothes, while samples taken from camera A and C share different clothes.
DeepChange~\cite{xu2023deepchange} is a large-scale long-term person re-id dataset containing a total number of 178,407 images from 1,121 IDs. Images are collected in diverse scenes and the time period spans across 12 months, thus the outfits of individuals are more varied. Note that DeepChange did not provide clothes annotations, so we alternatively employ the camera labels as clothes labels, following~\cite{gu2022clothes}. 

\textbf{Evaluation Protocols.} \ 
We adopt top-1 accuracy and mAP as evaluation metrics. Three evaluation settings are covered during testing: (1) \textbf{General setting}: all gallery samples are used to calculate accuracy; (2) \textbf{Clothes Changing setting (CC)}: only clothes-changing samples are used to calculate accuracy; (3) \textbf{Same Clothes setting (SC)}: only clothes-consistent samples are used to calculate accuracy. We mainly focus on the \textbf{Clothes Changing setting (CC)}, which is the most challenging setting of all three settings above. Following~\cite{gu2022clothes}, for LTCC, we report the accuracy of general and clothes-changing re-id. For PRCC, we report the accuracy of same-clothes and clothes-changing re-id. For DeepChange, we report the accuracy of general setting since the clothes are not labeled in test set.

\subsection{Implementation Details}
\label{sec: implementation details}
We adopt ResNet50~\cite{he2016deep} pretrained on ImageNet~\cite{deng2009imagenet} as backbone. Input images are resized to $384\times192$. Random horizontal flipping, random cropping and random erasing~\cite{zhong2020random} are used. We combine global max pooling and global average pooling to enrich information in the output feature, following~\cite{huang2021clothing}. A batch contains 64 images from 8 identities. Adam optimizer is utilized to train the model for 60 epochs. The initial learning rate is set to $3.5\times 10^{-4}$, decreasing by a factor of 10 every 20 epochs. In the total loss function (Eq.~\ref{eq: total loss}) , $\alpha_{ir}$ and $\alpha_{re}$ are set to $0.5$. The margin $\lambda_{fv}$ in Eq.~\ref{eq: feature variance loss} is set to $1.0$.

\subsection{Comparison with State-of-the-art Methods}
\label{sec: comparison with sota}
We compare our proposed method with conventional clothes-consistent re-id methods~\cite{hou2019interaction, zhou2019omni} and clothes-changing re-id methods~\cite{yang2019person, qian2020long, hong2021fine, gu2022clothes, yang2023good, han2023clothing, jin2022cloth, chen2021deep, liu2023learning, cui2023dcr, liu2023clothes, huang2024meta, guo2023semantic}. Our baseline is conducted by using an ImageNet-pretrained ResNet50 model as feature extractor, training with classification loss and triplet loss. As shown in Tab.~\ref{tab: comparison with sota}, our method outperforms the baseline as well as current state-of-the-art methods on three widely used re-id benchmarks, proving the superiority of FAIM. Notably, compared to methods ~\cite{yang2019person, qian2020long, hong2021fine, jin2022cloth, chen2021deep, cui2023dcr, liu2023learning}, FAIM only requires RGB modality, which is computationally efficient and eliminates the prediction errors associated with auxiliary modalities. A slight disadvantage to CCFA~\cite{han2023clothing} and SCNet~\cite{guo2023semantic} in the `CC' setting of PRCC is because CCFA additionally adopts a generative feature augmentation strategy, while SCNet leverages extra human silhouette modality. 

\begin{table*}[tbp]
\centering
\caption{Ablation study on the effectiveness of each component of FAIM. `FD', `IM' and `IBFW' denote Feature Decoupling, Intermediary Matching and Intermediary-Based Feasibility Weighting modules, respectively. `RR' denotes conventional re-ranking methods. `k-r' and `GNN' denote results of employing re-ranking with \textit{k-reciprocal}~\cite{zhong2017re} and \textit{GNN-based}~\cite{zhang2020understanding} methods, respectively. All the methods are tested on the clothes-changing (CC) setting of all benchmarks.}
\scalebox{1.1}{
\begin{tabular}{cccc|cc|cc|cc|cc|cc|cc}
\hline
    \multirow{3}*{FD} & \multirow{3}*{RR} & \multirow{3}*{IM} & \multirow{3}*{IBFW} & \multicolumn{4}{c|}{LTCC} & \multicolumn{4}{c|}{PRCC} & \multicolumn{4}{c}{DeepChange} \\
    \cline{5-16} &&&& \multicolumn{2}{c|}{k-r} & \multicolumn{2}{c|}{GNN} & \multicolumn{2}{c|}{k-r} & \multicolumn{2}{c|}{GNN} & \multicolumn{2}{c|}{k-r} & \multicolumn{2}{c}{GNN} \\
    \cline{5-16} &&&& top-1 & mAP & top-1 & mAP & top-1 & mAP & top-1 & mAP & top-1 & mAP & top-1 & mAP \\
\hline
    \XSolidBrush & \XSolidBrush & \XSolidBrush & \XSolidBrush & 36.0 & 16.0 & 36.0 & 16.0 & 53.4 & 53.8 & 53.4 & 53.8 & 53.8 & 18.1 & 53.8 & 18.1 \\
    \Checkmark & \XSolidBrush & \XSolidBrush & \XSolidBrush & 39.8 & 17.2 & 39.8 & 17.2 & 54.7 & 55.1 & 54.7 & 55.1 & 55.0 & 18.9 & 55.0 & 18.9 \\
    \Checkmark & \Checkmark & \XSolidBrush & \XSolidBrush & 43.6 & 23.1 & 42.1 & 20.9 & 55.3 & 57.5 & 55.0 & 58.6 & 57.1 & 24.5 & 55.9 & 23.7 \\
    \Checkmark & \XSolidBrush & \Checkmark & \XSolidBrush & 44.9 & 23.9 & 45.2 & 24.2 & 57.3 & 57.9 & 56.0 & 59.7 & 60.1 & 25.9 & 56.9 & 24.0 \\
    \hline
    \Checkmark & \XSolidBrush & \Checkmark & \Checkmark & \textbf{48.2} & \textbf{27.5} & \textbf{46.2} & \textbf{26.0} & \textbf{60.9} & \textbf{62.0} & \textbf{59.8} & \textbf{62.5} & \textbf{61.5} & \textbf{28.7} & \textbf{58.9} & \textbf{26.6} \\
\hline
\end{tabular}}
\label{tab: component ablation}
\end{table*}

\begin{table}[!htbp]
\centering
\caption{Ablation study on the effectiveness of each matching route ($A, B, C$) in IM module. Top-1 and mAP of the clothes-changing (CC) setting is reported.}
\begin{tabular}{ccc|cc|cc}
\hline
    \multirow{2}*{A} & \multirow{2}*{B} & \multirow{2}*{C} & \multicolumn{2}{c|}{LTCC} & \multicolumn{2}{c}{PRCC} \\
    \cline{4-7} &&& top-1 & mAP & top-1 & mAP \\
\hline
     &  &  & 43.6 & 23.1 & 55.3 & 57.5 \\
\hline
    \Checkmark &  &  & 45.7 & 26.1 & 57.6 & 61.1 \\
     & \Checkmark &  & 47.2 & 26.0 & 58.8 & 60.8 \\
     &  & \Checkmark & 44.6 & 23.2 & 59.4 & 60.1 \\
\hline
    \Checkmark & \Checkmark &  & 47.0 & 27.2 & 58.3 & 61.7  \\
    \Checkmark &  & \Checkmark & 46.2 & 26.1 & 59.6 & 61.7 \\
     & \Checkmark & \Checkmark & 48.2 & 25.6 & 58.8 & 61.1 \\
\hline
    \Checkmark & \Checkmark & \Checkmark & \textbf{48.2} & \textbf{27.5} & \textbf{60.9} & \textbf{62.0} \\
\hline
\end{tabular}
\label{tab:route ablation}
\end{table}

\begin{table}[tbp]
\centering
\caption{Ablation on the effectiveness of IBFW. `w/o SC' denotes same-clothes samples for intermediary matching are excluded. `-50\% high reliability' denotes that the samples with top 50\% highest reliability are excluded for intermediary matching. `all' denotes no samples are excluded. `w/o IM' denotes conventional re-ranking~\cite{zhong2017re} without our proposed IM. Top-1 and mAP of the clothes-changing (CC) setting is reported.}
\scalebox{1.0}{
\begin{tabular}{l|c|cc|cc}
\hline
    \multicolumn{2}{c|}{\multirow{2}{*}{method}} & \multicolumn{2}{c|}{LTCC} & \multicolumn{2}{c}{PRCC} \\
    \cline{3-6} \multicolumn{2}{c|}{} & top-1 & mAP & top-1 & mAP \\
\hline
     \multicolumn{2}{c|}{w/o IM} & 43.6 & 23.1 & 55.3 & 57.5 \\
\hline
    \multirow{3}*{w/o IBFW} & w/o SC & 43.1 & 22.0 & 57.1 & 60.6 \\
     & -50\% high reliability & 37.8 & 20.7 & 56.6 & 60.2  \\
     & all & 44.9 & 23.9 & 57.3 & 57.9  \\
\hline
    \multirow{3}*{\makecell{w/ IBFW \\ (\textbf{FAIM})}} & w/o SC & 45.4 & 25.1 & 57.3 & 61.2  \\
     & -50\% high reliability & 44.1 & 22.5 & 57.3 & 60.1  \\
     & all & 48.2 & 27.5 & 60.9 & 62.0 \\
\hline
\end{tabular}}
\label{tab: intermediary-based feasibility re-weighting ablation}
\end{table}

\subsection{Ablation Study}
\label{sec: ablation study}
\textbf{The effectiveness of FAIM components}. \
Tab.~\ref{tab: component ablation} represents the results of ablation study on major components of FAIM. As shown in Tab.~\ref{tab: component ablation}, on LTCC, PRCC and DeepChange benchmarks, FAIM consistently outperforms the baseline in both k-r and GNN algorithms. Moreover, by jointly using clothes-relevant and clothes-irrelevant features, the IM module brings substantial performance gain compared to re-ranking methods~\cite{zhong2017re, zhang2020understanding} only using clothes-irrelevant features (RR). The results highlight the superiority of IM over conventional re-ranking algorithms. Additionally, the IBFW module brings an obvious boost to the intermediary matching process, which verifies the effectiveness of incorporating feasibility-awareness into IM module. 

\begin{figure}[tbp]
    \centering
    \begin{subfigure}
        \centering
        \includegraphics[width=0.65\linewidth]{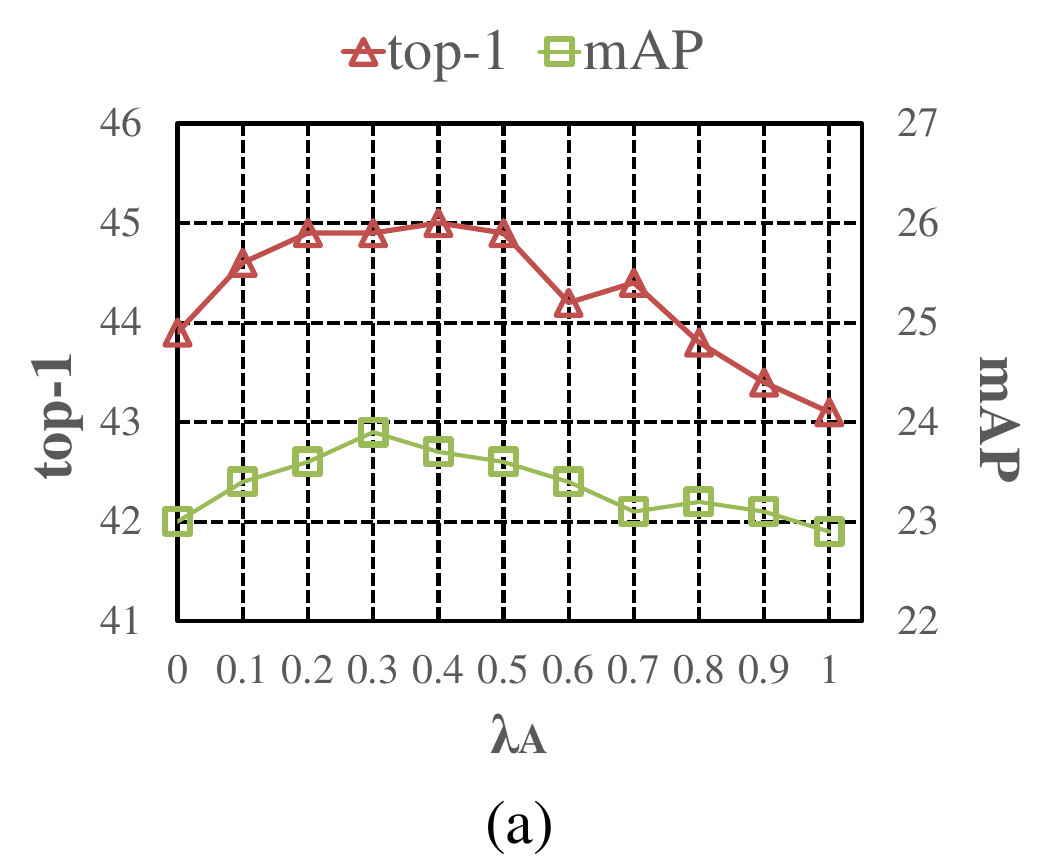}
        \label{fig:lambda_a}
    \end{subfigure}
    \begin{subfigure}
        \centering
        \includegraphics[width=0.65\linewidth]{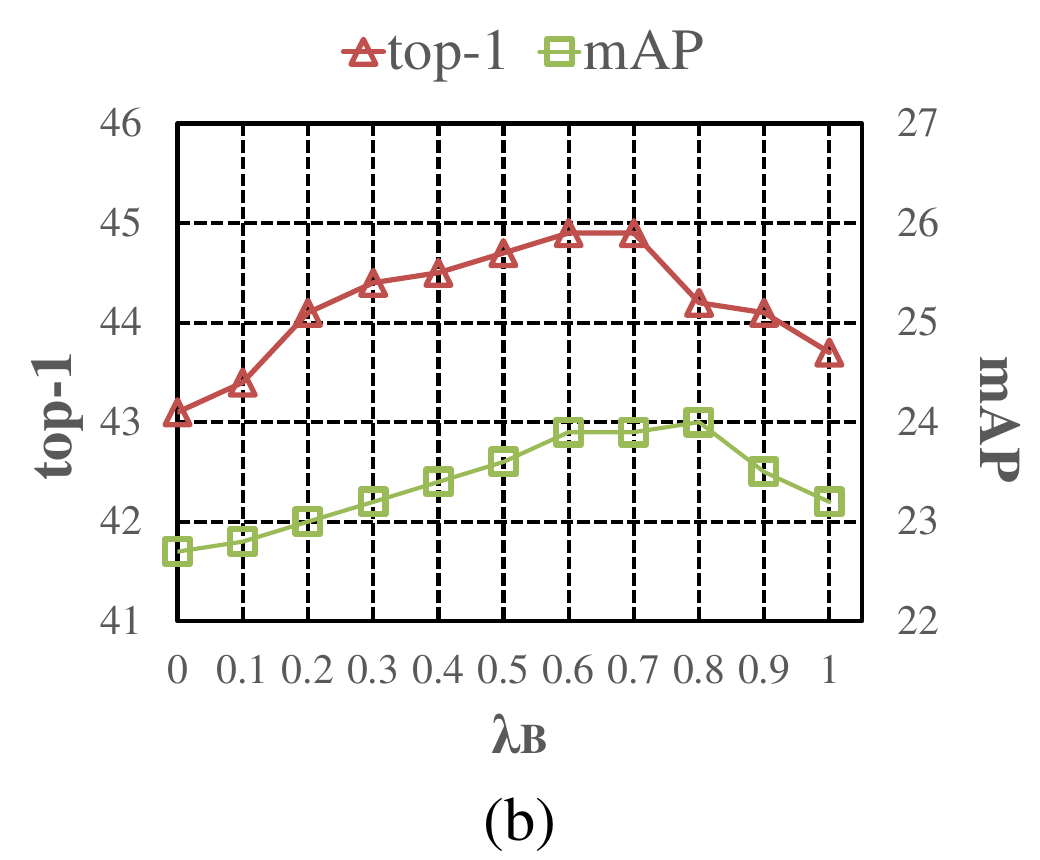}
        \label{fig:lambda_b}
    \end{subfigure}
    \begin{subfigure}
        \centering
        \includegraphics[width=0.65\linewidth]{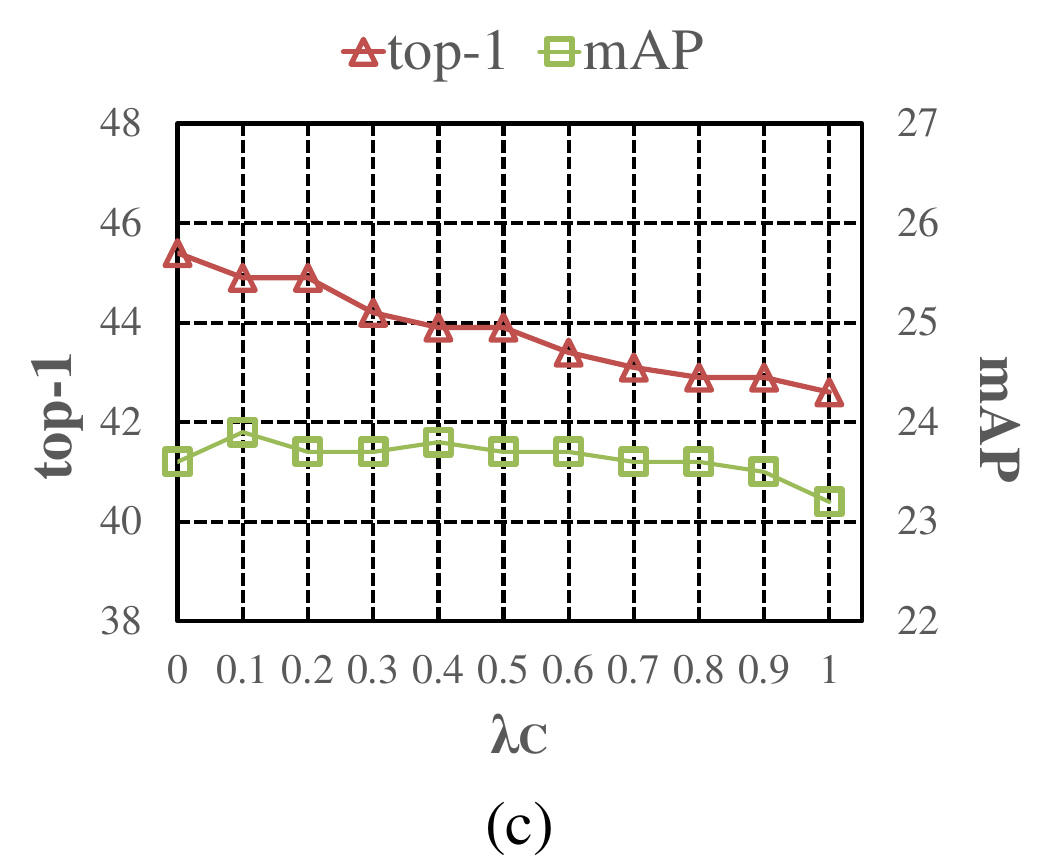}
        \label{fig:lambda_c}
    \end{subfigure}
    \caption{The top-1 accuracy and mAP with different $s_A$, $s_B$ and $s_C$ on the clothes-changing setting of LTCC. In (a) , we fix $s_B$ to $0.6$ and $s_C$ to $0.1$. In (b) , we fix $s_A$ to $0.3$ and $sa_C$ to $0.1$. In (c) , we fix $s_A$ to $0.3$ and $s_B$ to $0.6$.}
    \label{fig: hyper-parameter}
\end{figure}

For the experiment of IM without IBFW (the fourth row), instead of dynamically assigning the feasibility scores $s_{\mathcal{A}} \sim s_{\mathcal{C}}$ for route $\mathcal{A} \sim \mathcal{C}$ (see Eq.~\ref{eq: final distance}), we set $s_{\mathcal{A}}$, $s_{\mathcal{B}}$ and $s_{\mathcal{C}}$ as fixed hyper-parameters. To determine the best value combination of $s_{\mathcal{A}}$, $s_{\mathcal{B}}$ and $s_{\mathcal{C}}$, we conduct grid search on these three hyper-parameters. Specifically, we utilize cross-validation method by randomly splitting $1/10$ identities from the training set as validation samples. We vary the values of $s_{\mathcal{A}}$, $s_{\mathcal{B}}$ and $s_{\mathcal{C}}$ from $0.0$ to $1.0$ by the step size of $0.1$, and the best combination of $s_{\mathcal{A}}$, $s_{\mathcal{B}}$ and $s_{\mathcal{C}}$ values are selected by evaluating the top-1 and mAP under clothes-changing setting (CC). Results are shown in Fig.~\ref{fig: hyper-parameter}. The performance is reported by varying the hyper-parameter denoted in the abscissa, while fixing the other two hyper-parameters. According to the results in Fig.~\ref{fig: hyper-parameter}, we select the hyper-parameter values at $s_{\mathcal{A}}=0.3$, $s_{\mathcal{B}}=0.6$ and $s_{\mathcal{C}}=0.1$ for the setting of IM without IBFW.

It is also worth noting that in Fig.~\ref{fig: hyper-parameter}(c), when not applying the IBFW, the performance of route $\mathcal{C}$ declines as the weight increases. This phenomenon can be explained as follows: Compared to route $\mathcal{A}$ and $\mathcal{B}$ which only involve one intermediary, route $\mathcal{C}$ involves two intermediaries. This increases the chance of introducing intermediaries with low availability or reliability, which downgrade the re-id accuracy. When not applying IBFW, the feasibility score  $s_{\mathcal{C}}$ of is fixed. Therefore, when encountered with intermediaries with low availability or low reliability, route $s_{\mathcal{C}}$ cannot be down-weighted accordingly to avoid the negative influence of low-quality intermediaries. As a result, route $s_{\mathcal{C}}$ exhibits performance decrease in Fig.~\ref{fig: hyper-parameter}(c).

\textbf{The effectiveness of matching routes in Intermediary Matching (IM) module}. \
In IM module, we designed three matching routes $s_{\mathcal{A}} \sim s_{\mathcal{C}}$ to address different circumstances that need intermediaries to facilitate matching. Tab.~\ref{tab:route ablation} conducts ablations to validate the effectiveness of each route. As shown in Tab.~\ref{tab:route ablation}, (1) Compared to the method not leveraging intermediary matching (the first row), utilizing any of the three routes individually could lead to performance gain. (2) Compared to using all three routes, only deploying one or two routes will cause performance drop. These results show the effectiveness of each matching route, and consolidate the importance of incorporating three matching routes altogether for the IM process.

\textbf{The effectiveness of Intermediary-Based Feasibility Re-weighting (IBFW) module.} \
In Tab.~\ref{tab: intermediary-based feasibility re-weighting ablation}, we conduct experiments to validate the effects of our IBFW module. Since IBFW is primarily designed to deal with situations where high-quality intermediaries are inaccessible, we manually simulate these situations and test the performance of IBFW: (1) To simulate situations of low intermediary availability, we create `w/o SC' setting by excluding same-clothes samples when fetching intermediaries with clothes-relevant features. (2) To simulate situations of low clothes-irrelevant identity information reliability, we create `-50\% high reliability' setting by excluding samples of the top 50\% highest reliability score $r^{id}$ from the gallery. As shown in Tab.~\ref{tab: intermediary-based feasibility re-weighting ablation}, under both situations, FAIM (`w/ IBFW') performs better than not using IBFW (`w/o IBFW'). By employing IBFW, the performance is less impaired in low availability or low reliability situations, and the advantage against conventional re-ranking methods (`w/o IM') can be maintained. Conclusively, IBFW is useful in addressing low-quality intermediaries and guaranteeing the robustness of IM against adverse data conditions.

\begin{table}[tbp]
\centering
\caption{Ablation study on the effectiveness of IIR module. `w/o IIR denotes training the model without adopting the IIR module, and testing without incorporating the reliability score. Top-1 and mAP of the clothes-changing (CC) setting is reported.}
\scalebox{1.0}{
\begin{tabular}{l|cc|cc}
\hline
    \multirow{2}*{method} & \multicolumn{2}{c}{LTCC} & \multicolumn{2}{c}{PRCC} \\
    \cline{2-5} & top-1 & mAP & top-1 & mAP \\
\hline
    w/o IIR & 45.7 & 24.0 & 57.6 & 61.5 \\
    \textbf{FAIM} & \textbf{48.2} & \textbf{27.5} & \textbf{60.9} & \textbf{62.0} \\
\hline
\end{tabular}
\label{tab:identity reliability block ablation}
}
\end{table}

\begin{table}[tbp]
\centering
\caption{Ablation study on the construction method of $\boldsymbol{\Sigma}^{ir}$ in the Identity Information Reliability block. top-1 and mAP of the clothes-changing (CC) setting on LTCC is reported.}
\scalebox{1.0}{
\begin{tabular}{c|cc}
\hline
    \multirow{2}*{$\boldsymbol{\Sigma}^{ir}$ construction} & \multicolumn{2}{c}{LTCC} \\
    \cline{2-3} & top-1 & mAP \\
\hline
    direct predict~\cite{yu2019robust} & 45.4 & 26.3 \\
    global $\boldsymbol{\Sigma}^{c}$~\cite{han2023clothing} & 44.6 & 26.2 \\
    instance-wise $\boldsymbol{\Sigma}^{c}$ (ours) & \textbf{48.2} & \textbf{27.5} \\
\hline
\end{tabular}}
\label{tab:identity variance construction ablation}
\end{table}

\begin{figure}[!htbp]
    \centering
    \includegraphics[width=1.0\linewidth]{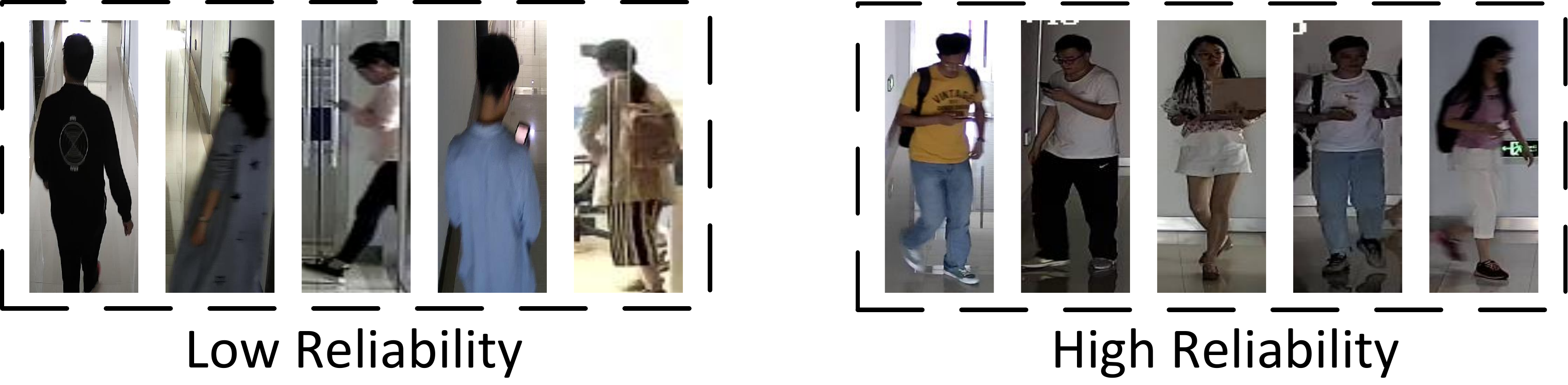}
    \caption{Visualization results of person with low and high reliability score of clothes-irrelevant identity information ($r^{id}$). The left side shows samples with $r^{id} < 0.5$, while the right side shows samples with $r^{id} > 0.5$.}
    \label{fig: identity reliability visualization}
\end{figure}

\textbf{The effectiveness of Identity Information Reliability (IIR) module.} \
In Tab.~\ref{tab:identity reliability block ablation}, we ablate on the effectiveness of our proposed IIR module. As the results in Tab.~\ref{tab:identity reliability block ablation} shows, the adoption of IIR brings a top-1 performance gain of $2.5\%$ and $3.3\%$ on LTCC and PRCC datasets, respectively. This verifies the importance of modeling the reliability of clothes-irrelevant information in our FAIM framework.

To give a more straightforward illustration on the function of reliability modeling, Fig.~\ref{fig: identity reliability visualization} visualizes some examples of pedestrian images with low and high reliability of clothes-irrelevant identity information. As shown in Fig.~\ref{fig: identity reliability visualization}, images with low reliability typically lack inadequate identity information, \textit{e.g.}, absence of frontal face view or incomplete body shape. In contrast, images with high reliability typically contain sufficient identity-related information. Notably, images on the left side get low reliability score even if they have clear clothes representation, which verifies that our IIR module can measure the reliability of identity information that is \textbf{\textit{clothes-irrelevant}}. This capability is crucial in clothes-changing scenarios, where only intermediaries with reliable \textbf{\textit{clothes-irrelevant}} identity cues can have high accuracy when matching to clothes-changing targets. 

Furthermore, results in Tab.~\ref{tab:identity variance construction ablation} support the aforementioned claim. As seen, the final performance when using clothes-changing variance $\boldsymbol{\Sigma}^{c}$ to construct $\boldsymbol{\Sigma}^{ir}$ is better than directly predicting the variance following~\cite{yu2019robust} which is unaware of clothes changes. Moreover, as shown in Tab.~\ref{tab:identity variance construction ablation}, our approach of constructing $\boldsymbol{\Sigma}^{c}$ in instance-wise manner is advantageous compared to using a global $\boldsymbol{\Sigma}^{c}$~\cite{han2023clothing} for every instance. The main reason is that different samples exhibit variations in their clothes-changing semantic directions, so that instance-specific clothes-changing variance could model the clothes-changing directions more precisely.

\begin{figure*}[h!t]
    \centering
    \includegraphics[width=0.85\linewidth]{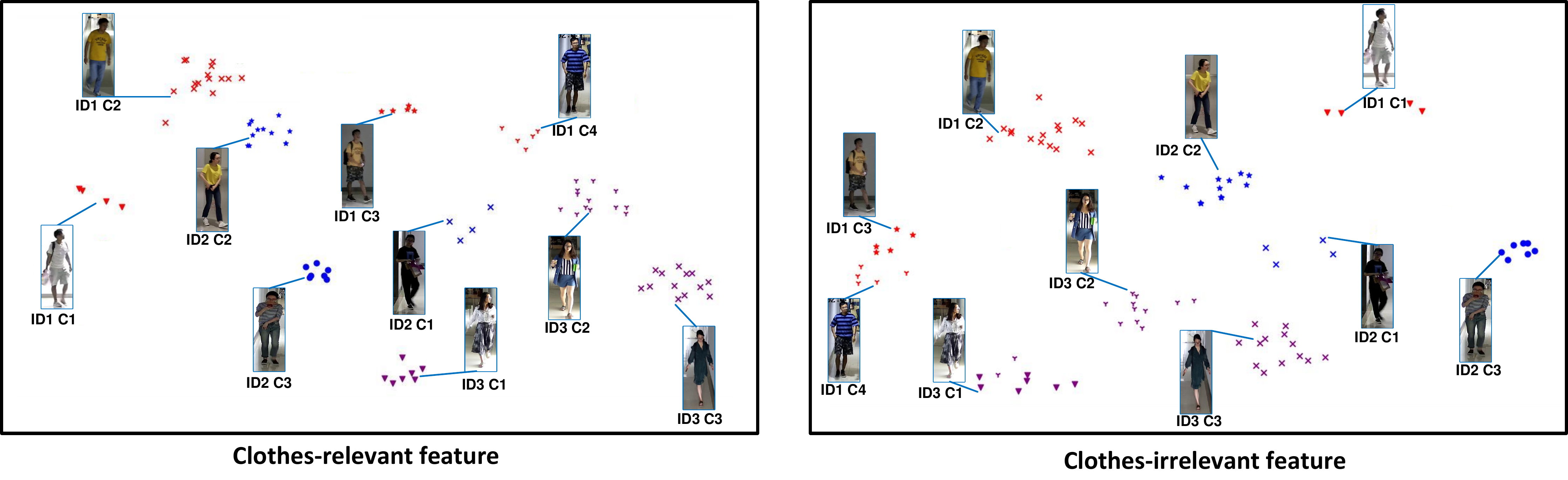}
    \caption{T-SNE~\cite{van2008visualizing} visualization results clothes-relevant feature and clothes-irrelevant feature derived from the FD module. Each color represents one identity, and each marker style stands for each type of clothing within each identity. The identity index and clothes index within each identity are labeled in the graph.}
    \label{fig: feature decoupling tsne visualization}
\end{figure*}

\begin{table}[tbp]
\centering
\caption{Ablation on the incorporation design of clothes-relevant and clothes-irrelevant features. We try to incorporate $f^{ir}$ and $f^{re}$ by concatenation along the channel dimension. `Baseline + Concat.' denotes using concatenated features to perform regular re-ranking~\cite{zhong2017re}. `FAIM + Concat.' denotes using concatenated features to substitute $f^{ir}$ and $f^{re}$, and perform the intermediary matching process. For `general', `SC' and `CC' settings, we report the Top-1 accuracy.}
\scalebox{1.0}{
\begin{tabular}{c|cc|cc}
\hline
    \multirow{2}{*}{method} & \multicolumn{2}{c|}{LTCC} & \multicolumn{2}{c}{PRCC} \\
    \cline{2-5} & general & CC & SC & CC \\
\hline
    Baseline & 73.4 & 36.0 & 100 & 53.4 \\
    Baseline + Concat. & 80.5 & 44.6 & 100 & 48.7 \\
    FAIM + Concat. & 78.7 & 47.1 & 100 & 55.1 \\
\hline
    FAIM & 79.5 & 48.2 & 100 & 60.9 \\
\hline
\end{tabular}}
\label{tab:incorporation ablation}
\end{table}

\textbf{Incorporation design of clothes-relevant and clothes-irrelevant features.} \
In FAIM, we design an intermediary matching pipeline which leverages clothes-relevant features $f^{re}$ and clothes-irrelevant features $f^{ir}$ to perform indirect matching routes. As defined in Eq.~\ref{eq: intermediary matching routes}, in every matching step, we selectively use $f^{ir}$ or $f^{re}$, but never jointly use both features in one step. To validate the advantage of our selective feature incorporation approach, we compare FAIM against jointly use both features in every matching step. When applying the joint usage approach, we substitute every feature in Eq.~\ref{eq: intermediary matching routes} with the concatenation of $f^{ir}$ and $f^{re}$. Results in Tab.~\ref{tab:incorporation ablation} shows that using the concatenation $f^{ir}$ and $f^{re}$ as the feature representation in FAIM (FAIM + Concat.) degrades the performance of FAIM on both LTCC and PRCC datasets, and the superiority of our selective feature incorporation design in FAIM is verified. Moreover, FAIM largely outcompetes the approach of concatenating $f^{ir}$ and $f^{re}$ and use the incorporated feature to perform direct matching (Baseline + Concat.), which proves that our carefully designed intermediary matching routes provide a more proper way to jointly utilize both clothes-relevant and clothes-irrelevant features.

\textbf{Effects of Introducing Clothes-relevant Features.} \
In the Intermediary Matching (IM) process, we introduce clothes-relevant features to help match more informative intermediaries. To ensure that clothes-relevant features have enough discrimination between different identities in the IM process, in Tab.~\ref{tab:clothes relevant feature ablation}, we calculate the portion of same-identity samples in the intermediaries when matching with clothes-relevant features, and compared with the method of using original feature $\boldsymbol{f}^{o}$ in the entire IM process (denoted as w/o $\boldsymbol{f}^{re}$). As shown in Tab.~\ref{tab:clothes relevant feature ablation}, in FAIM method (and each of the matching routes $\mathcal{A} \sim \mathcal{C}$), the portion of same-identity intermediaries when matching with clothes-relevant features are comparable with using the original re-id feature. Moreover, we calculated the average feasibility score of same-identity intermediaries ($s_{Pos}$) and different-identity $s_{Neg}$ intermediaries, respectively. From Tab.~\ref{tab:clothes relevant feature ablation}, we can observe that the feasibility scores of same-identity samples are higher than those of different-identity samples. Therefore, we can conclude that our clothes-relevant feature representation itself could discriminate different identities, thanks to the identity-discriminative supervision signals. Meanwhile, our IBFW module can further eliminate the effects of different-identity samples, by assigning low feasibility scores to them, thereby decreasing the weights of these samples in the whole matching process.

\begin{table}[tbp]
\centering
\caption{Ablation study on the effects of clothes-relevant features in IM process. `Pos.ID(\%)' stands for the portion of intermediaries with the same ID as query, when matching with clothes-relevant feature. `$s_{Pos}$' and `$s_{Neg}$' indicates the average feasibility score of same-identity and different-identity intermediaries, respectively. `w/o $\boldsymbol{f}^{re}$' stands for using original feature $\boldsymbol{f}^{o}$ instead of clothes-relevant feature $\boldsymbol{f}^{re}$ for the entire IM process.}
\scalebox{1.0}{
\begin{tabular}{c|ccc|ccc}
\hline
    \multirow{2}*{method} & \multicolumn{3}{c}{LTCC} & \multicolumn{3}{|c}{PRCC} \\
    \cline{2-7} & Pos.ID(\%) & $s_{Pos}$ & $s_{Neg}$ & Pos.ID(\%) & $s_{Pos}$ & $s_{Neg}$ \\
\hline
    w/o $\boldsymbol{f}^{re}$ & 79.2 & - & - & 62.1 & - & - \\
\hline
    Route $\mathcal{A}$ & 82.6 & 0.50 & 0.09 & 61.6 & 0.48 & 0.26 \\
    Route $\mathcal{B}$ & 79.2 & 0.47 & 0.12 & 61.8 & 0.48 & 0.28 \\
    Route $\mathcal{C}$ & 78.2 & 0.24 & 0.07 & 58.4 & 0.22 & 0.15 \\
\hline
    FAIM & 80.0 & 0.40 & 0.09 & 60.6 & 0.39 & 0.23 \\
\hline
\end{tabular}
\label{tab:clothes relevant feature ablation}
}
\end{table}

\textbf{Hyper-parameter analysis.} \
In our training loss function $\mathcal{L}$ (see Eq.~\ref{eq: total loss}), there are two hyper-parameters, $\alpha^{ir}$ and $\alpha^{re}$. We set both of them as $0.5$. Here we conduct an hyper-parameter analysis on these two coefficients. The results are shown in Tab.~\ref{tab:hyperparameter ir re}. Under our selection of $\alpha^{ir}$ and $\alpha^{re}$, the model gains the overall best performance on both LTCC and PRCC benchmarks.

\begin{figure*}[tbp]
    \centering
    \includegraphics[width=0.85\linewidth]{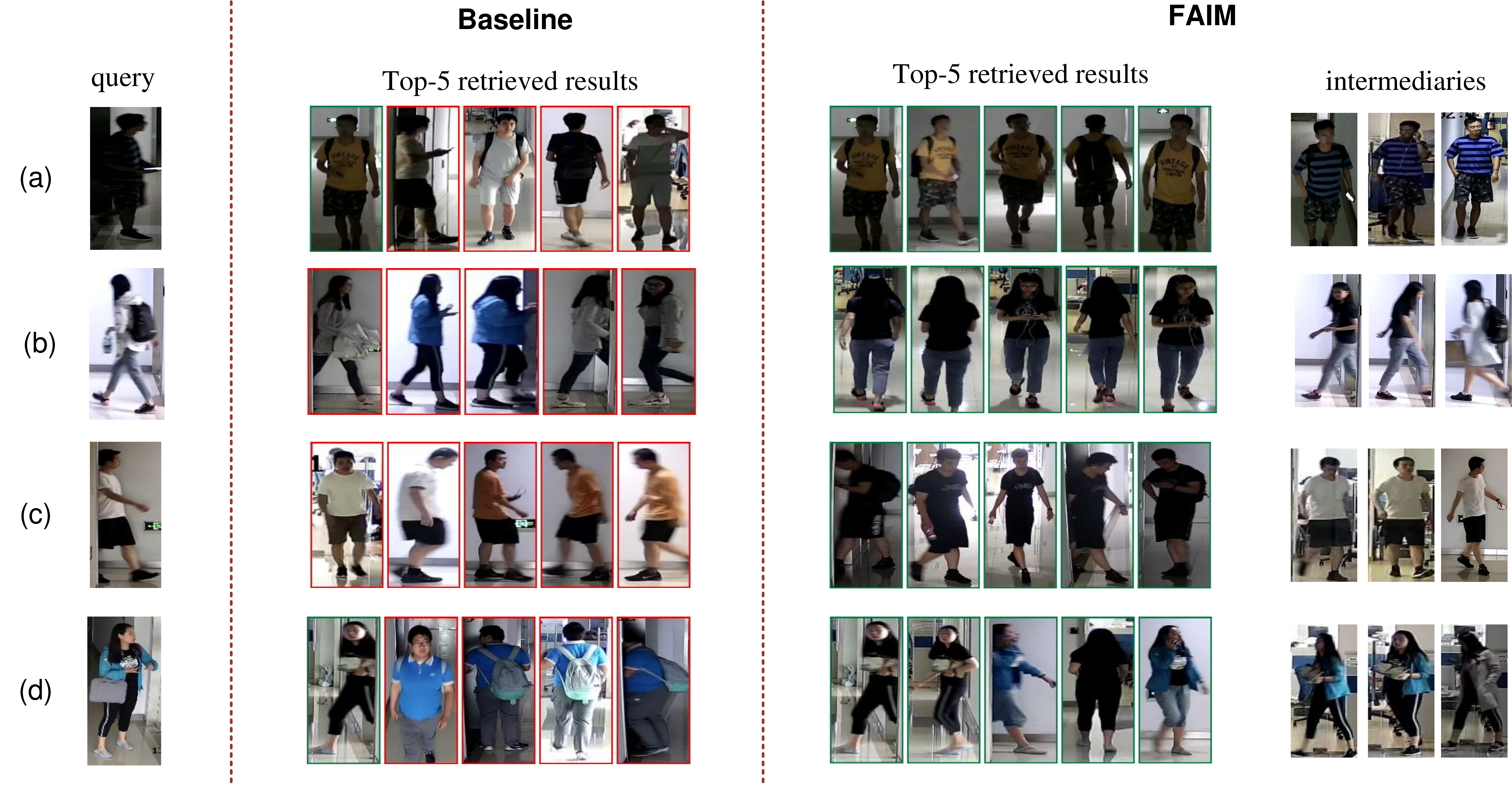}
    \caption{Visualization of FAIM matching examples. For each query, the top-5 retrievals of baseline and FAIM methods are shown. Corresponding intermediary samples in the FAIM matching process are also shown. The green box denotes positive matches that share the same identity as query, while red box denotes negative matches.}
    \label{fig: matching result visualization}
\end{figure*}

\begin{table}[tbp]
\centering
\caption{Hyper-parameter analysis. To verify the optimality of our hyper-parameter selection (setting both $\alpha^{ir}$ and $\alpha^{re}$ at $0.5$), we fix $\alpha^{re}$ at $0.5$ and change $\alpha^{ir}$, as well as fixing $\alpha^{re}$ at $0.5$ and change $\alpha^{re}$. We report the Top-1 and mAP results of clothes-changing (CC) setting on all datasets.}
\scalebox{1.0}{
\begin{tabular}{cc|cc|cc}
\hline
    \multirow{2}{*}{$\alpha_{ir}$} & \multirow{2}{*}{$\alpha_{re}$} & \multicolumn{2}{c|}{LTCC} & \multicolumn{2}{c}{PRCC} \\
    \cline{3-6} && top-1 & mAP & top-1 & mAP \\
\hline
    0.1 & 0.5 & 47.2 & 26.5 & 53.9 & 57.0 \\
    0.3 & 0.5 & 46.7 & 26.9 & 57.4 & 59.5 \\
    \textbf{0.5} & \textbf{0.5} & 
    \textbf{48.2} & \textbf{27.5} & \textbf{60.9} & \textbf{62.0} \\
    0.7 & 0.5 & 47.4 & 27.1 & 59.0 & 60.8 \\
    0.9 & 0.5 & 46.4 & 24.8 & 56.7 & 59.3 \\
\hline
    0.5 & 0.1 & 45.7 & 26.4 & 53.0 & 58.4 \\
    0.5 & 0.3 & 47.2 & 27.5 & 57.7 & 61.5 \\
    \textbf{0.5} & \textbf{0.5} & \textbf{48.2} & \textbf{27.5} & \textbf{60.9} & \textbf{62.0} \\
    0.5 & 0.7 & \textbf{48.2} & 27.3 & 59.9 & 61.0 \\
    0.5 & 0.9 & 44.6 & 26.9 & 57.1 & 61.8 \\
\hline
\end{tabular}}
\label{tab:hyperparameter ir re}
\end{table}

\subsection{Visualization}
\textbf{Visualization of Feature Decoupling Results.} \
Fig.~\ref{fig: feature decoupling tsne visualization} uses t-SNE~\cite{van2008visualizing} tool to visualize the sample distribution of the clothes-relevant and clothes-irrelevant feature spaces. From Fig.~\ref{fig: feature decoupling tsne visualization}, we can observe that: (1) Compared with clothes-irrelevant features, samples of the same clothing exhibit a more compact clustering in clothes-relevant feature space (\textit{e.g.}, the features of ID3-C2). Therefore, by utilizing clothes-relevant features in intermediary matching, we can retrieve same-clothes intermediaries of the same identity more easily than using clothes-irrelevant features. This efficiency arises from the reduced presence of outliers within each clothes cluster. (2) In the clothes-relevant feature space, samples of the same identity remain gathered rather than being split up by samples of other identities. Also, for samples from different identities with similar clothes (\textit{e.g.}, ID1-C2 and ID2-C2), there is a clear discrepancy between the features of different identities. This indicates that our clothes label and special designed loss function for learning clothes-relevant features can preserve the identity-related information while mining fine-grained clothes cues. Therefore, by using our clothes-relevant feature for intermediary matching, we can ensure matching same-identity intermediaries in priority, and preclude noisy intermediaries with same clothes but different identities as query. In summary, our feature decoupling method adeptly generates clothes-relevant and clothes-irrelevant features suitable for intermediary matching.

\textbf{Visualization of Re-identification Results.} \
Fig.~\ref{fig: matching result visualization} visualizes some re-id results of FAIM under clothes-changing setting. We present the top-5 matching results of baseline and FAIM, as well as the top intermediaries fetched during the IM process of FAIM. As shown Fig.~\ref{fig: matching result visualization}, for the query image lacking clothes-irrelevant identity clues (ambiguous body shape in (a), and absence of facial view in (b)), FAIM can match it to positive targets via intermediaries with adequate clothes-irrelevant information, such as clear body shape and facial views. In (c) and (d), we can see that FAIM can match the query and targets with large clothes-irrelevant intra-class variance (pose, view or body shape), by utilizing intermediates sharing aligned body shape with query/targets. These results show that FAIM can effectively leverage intermediaries with richer identity-related characteristics to perform person retrieval.

\section{Conclusion}
In this paper, we propose a Feasibility-Aware Intermediary Matching (FAIM) framework. In our framework, the Intermediary Matching (IM) module employs intermediate samples to build a novel multiple-route intermediary-assisted re-id process. By jointly leveraging clothes-relevant and clothes-irrelevant features, IM can effectively match samples that are hard to pair using clothes-irrelevant features alone. To address the challenge of varying intermediary sample quality, the Intermediary-Based Feasibility-Weighting (IBFW) module dynamically assigns feasibility weights to different intermediary matching routes according to the availability and reliability of intermediaries. Through the application of IBFW, the performance of FAIM can be maintained when desired intermediary samples are scarce. Extensive experiments demonstrate that our method can achieve state-of-the-art performance on mainstream clothes-changing re-id benchmarks.

\textbf{Broader impacts}.\quad The proposed method boosts the performance of clothes-changing re-id, making it more practical in security, intelligent monitoring, \textit{etc}. Meanwhile, the higher performance may raise the risk of privacy breaches, therefore the collection and usage of pedestrian data should be regulated.

\section*{Acknowledgements}
This work is partially supported by Natural Science Foundation of China (NSFC): 62306301, 62376259, 62276246, in part by Fundamental Research Funds for the Central Universities, and in part by the National Postdoctoral Program for Innovative Talents under Grant BX20220310.

{
    \bibliographystyle{IEEEtran}
    \bibliography{ref}

\begin{thebibliography}{10}
\providecommand{\url}[1]{#1}
\csname url@samestyle\endcsname
\providecommand{\newblock}{\relax}
\providecommand{\bibinfo}[2]{#2}
\providecommand{\BIBentrySTDinterwordspacing}{\spaceskip=0pt\relax}
\providecommand{\BIBentryALTinterwordstretchfactor}{4}
\providecommand{\BIBentryALTinterwordspacing}{\spaceskip=\fontdimen2\font plus
\BIBentryALTinterwordstretchfactor\fontdimen3\font minus \fontdimen4\font\relax}
\providecommand{\BIBforeignlanguage}[2]{{%
\expandafter\ifx\csname l@#1\endcsname\relax
\typeout{** WARNING: IEEEtran.bst: No hyphenation pattern has been}%
\typeout{** loaded for the language `#1'. Using the pattern for}%
\typeout{** the default language instead.}%
\else
\language=\csname l@#1\endcsname
\fi
#2}}
\providecommand{\BIBdecl}{\relax}
\BIBdecl

\bibitem{hou2020iaunet}
R.~Hou, B.~Ma, H.~Chang, X.~Gu, S.~Shan, and X.~Chen, ``Iaunet: Global context-aware feature learning for person reidentification,'' \emph{IEEE TNNLS}, vol.~32, no.~10, pp. 4460--4474, 2020.

\bibitem{gu2019temporal}
X.~Gu, B.~Ma, H.~Chang, S.~Shan, and X.~Chen, ``Temporal knowledge propagation for image-to-video person re-identification,'' in \emph{ICCV}, 2019.

\bibitem{sun2018beyond}
Y.~Sun, L.~Zheng, Y.~Yang, Q.~Tian, and S.~Wang, ``Beyond part models: Person retrieval with refined part pooling (and a strong convolutional baseline),'' in \emph{ECCV}, 2018.

\bibitem{gu2020appearance}
X.~Gu, H.~Chang, B.~Ma, H.~Zhang, and X.~Chen, ``Appearance-preserving 3d convolution for video-based person re-identification,'' in \emph{ECCV}, 2020.

\bibitem{hou2021bicnet}
R.~Hou, H.~Chang, B.~Ma, R.~Huang, and S.~Shan, ``Bicnet-tks: Learning efficient spatial-temporal representation for video person re-identification,'' in \emph{CVPR}, 2021.

\bibitem{bai2022salient}
S.~Bai, B.~Ma, H.~Chang, R.~Huang, and X.~Chen, ``Salient-to-broad transition for video person re-identification,'' in \emph{CVPR}, 2022.

\bibitem{wang2018learning}
G.~Wang, Y.~Yuan, X.~Chen, J.~Li, and X.~Zhou, ``Learning discriminative features with multiple granularities for person re-identification,'' in \emph{ACM MM}, 2018.

\bibitem{yang2019person}
Q.~Yang, A.~Wu, and W.-S. Zheng, ``Person re-identification by contour sketch under moderate clothing change,'' \emph{IEEE TPAMI}, vol.~43, no.~6, pp. 2029--2046, 2019.

\bibitem{hong2021fine}
P.~Hong, T.~Wu, A.~Wu, X.~Han, and W.-S. Zheng, ``Fine-grained shape-appearance mutual learning for cloth-changing person re-identification,'' in \emph{CVPR}, 2021.

\bibitem{qian2020long}
X.~Qian, W.~Wang, L.~Zhang, F.~Zhu, Y.~Fu, T.~Xiang, Y.-G. Jiang, and X.~Xue, ``Long-term cloth-changing person re-identification,'' in \emph{ACCV}, 2020.

\bibitem{zhang2020learning}
P.~Zhang, J.~Xu, Q.~Wu, Y.~Huang, and X.~Ben, ``Learning spatial-temporal representations over walking tracklet for long-term person re-identification in the wild,'' \emph{IEEE TMM}, vol.~23, pp. 3562--3576, 2020.

\bibitem{jin2022cloth}
X.~Jin, T.~He, K.~Zheng, Z.~Yin, X.~Shen, Z.~Huang, R.~Feng, J.~Huang, Z.~Chen, and X.-S. Hua, ``Cloth-changing person re-identification from a single image with gait prediction and regularization,'' in \emph{CVPR}, 2022.

\bibitem{gu2022clothes}
X.~Gu, H.~Chang, B.~Ma, S.~Bai, S.~Shan, and X.~Chen, ``Clothes-changing person re-identification with rgb modality only,'' in \emph{CVPR}, 2022.

\bibitem{huang2019beyond}
Y.~Huang, J.~Xu, Q.~Wu, Y.~Zhong, P.~Zhang, and Z.~Zhang, ``Beyond scalar neuron: Adopting vector-neuron capsules for long-term person re-identification,'' \emph{IEEE TCSVT}, vol.~30, no.~10, pp. 3459--3471, 2019.

\bibitem{huang2021clothing}
Y.~Huang, Q.~Wu, J.~Xu, Y.~Zhong, and Z.~Zhang, ``Clothing status awareness for long-term person re-identification,'' in \emph{ICCV}, 2021.

\bibitem{shu2021semantic}
X.~Shu, G.~Li, X.~Wang, W.~Ruan, and Q.~Tian, ``Semantic-guided pixel sampling for cloth-changing person re-identification,'' \emph{IEEE SPL}, vol.~28, pp. 1365--1369, 2021.

\bibitem{xu2023deepchange}
P.~Xu and X.~Zhu, ``Deepchange: A long-term person re-identification benchmark with clothes change,'' in \emph{ICCV}, 2023.

\bibitem{fan2020learning}
L.~Fan, T.~Li, R.~Fang, R.~Hristov, Y.~Yuan, and D.~Katabi, ``Learning longterm representations for person re-identification using radio signals,'' in \emph{CVPR}, 2020.

\bibitem{chen2021learning}
J.~Chen, X.~Jiang, F.~Wang, J.~Zhang, F.~Zheng, X.~Sun, and W.-S. Zheng, ``Learning 3d shape feature for texture-insensitive person re-identification,'' in \emph{CVPR}, 2021.

\bibitem{li2021diverse}
Y.~Li, J.~He, T.~Zhang, X.~Liu, Y.~Zhang, and F.~Wu, ``Diverse part discovery: Occluded person re-identification with part-aware transformer,'' in \emph{CVPR}, 2021.

\bibitem{chum2007total}
O.~Chum, J.~Philbin, J.~Sivic, M.~Isard, and A.~Zisserman, ``Total recall: Automatic query expansion with a generative feature model for object retrieval,'' in \emph{ICCV}, 2007.

\bibitem{li2012common}
W.~Li, Y.~Wu, M.~Mukunoki, and M.~Minoh, ``Common-near-neighbor analysis for person re-identification,'' in \emph{ICIP}, 2012.

\bibitem{ye2016person}
M.~Ye, C.~Liang, Y.~Yu, Z.~Wang, Q.~Leng, C.~Xiao, J.~Chen, and R.~Hu, ``Person reidentification via ranking aggregation of similarity pulling and dissimilarity pushing,'' \emph{IEEE TMM}, vol.~18, no.~12, pp. 2553--2566, 2016.

\bibitem{bai2019re}
S.~Bai, P.~Tang, P.~H. Torr, and L.~J. Latecki, ``Re-ranking via metric fusion for object retrieval and person re-identification,'' in \emph{CVPR}, 2019.

\bibitem{tan2021instance}
F.~Tan, J.~Yuan, and V.~Ordonez, ``Instance-level image retrieval using reranking transformers,'' in \emph{ICCV}, 2021.

\bibitem{qin2011hello}
D.~Qin, S.~Gammeter, L.~Bossard, T.~Quack, and L.~Van~Gool, ``Hello neighbor: Accurate object retrieval with k-reciprocal nearest neighbors,'' in \emph{CVPR}, 2011.

\bibitem{zhong2017re}
Z.~Zhong, L.~Zheng, D.~Cao, and S.~Li, ``Re-ranking person re-identification with k-reciprocal encoding,'' in \emph{CVPR}, 2017.

\bibitem{zhang2020understanding}
X.~Zhang, M.~Jiang, Z.~Zheng, X.~Tan, E.~Ding, and Y.~Yang, ``Understanding image retrieval re-ranking: a graph neural network perspective,'' \emph{arXiv preprint arXiv:2012.07620}, 2020.

\bibitem{gilmer2017neural}
J.~Gilmer, S.~S. Schoenholz, P.~F. Riley, O.~Vinyals, and G.~E. Dahl, ``Neural message passing for quantum chemistry,'' in \emph{ICML}, 2017.

\bibitem{hermans2017defense}
A.~Hermans, L.~Beyer, and B.~Leibe, ``In defense of the triplet loss for person re-identification,'' \emph{arXiv preprint arXiv:1703.07737}, 2017.

\bibitem{deng2009imagenet}
J.~Deng, W.~Dong, R.~Socher, L.-J. Li, K.~Li, and L.~Fei-Fei, ``Imagenet: A large-scale hierarchical image database,'' in \emph{CVPR}, 2009.

\bibitem{he2016deep}
K.~He, X.~Zhang, S.~Ren, and J.~Sun, ``Deep residual learning for image recognition,'' in \emph{CVPR}, 2016.

\bibitem{hou2019interaction}
R.~Hou, B.~Ma, H.~Chang, X.~Gu, S.~Shan, and X.~Chen, ``Interaction-and-aggregation network for person re-identification,'' in \emph{CVPR}, 2019.

\bibitem{zhou2019omni}
K.~Zhou, Y.~Yang, A.~Cavallaro, and T.~Xiang, ``Omni-scale feature learning for person re-identification,'' in \emph{ICCV}, 2019.

\bibitem{zhong2020random}
Z.~Zhong, L.~Zheng, G.~Kang, S.~Li, and Y.~Yang, ``Random erasing data augmentation,'' in \emph{AAAI}, 2020.

\bibitem{han2023clothing}
K.~Han, S.~Gong, Y.~Huang, L.~Wang, and T.~Tan, ``Clothing-change feature augmentation for person re-identification,'' in \emph{CVPR}, 2023.

\bibitem{yang2023good}
Z.~Yang, M.~Lin, X.~Zhong, Y.~Wu, and Z.~Wang, ``Good is bad: Causality inspired cloth-debiasing for cloth-changing person re-identification,'' in \emph{CVPR}, 2023.

\bibitem{dou2022reliability}
Z.~Dou, Z.~Wang, W.~Chen, Y.~Li, and S.~Wang, ``Reliability-aware prediction via uncertainty learning for person image retrieval,'' in \emph{ECCV}, 2022.

\bibitem{yu2019robust}
T.~Yu, D.~Li, Y.~Yang, T.~M. Hospedales, and T.~Xiang, ``Robust person re-identification by modelling feature uncertainty,'' in \emph{ICCV}, 2019.

\bibitem{wang2019implicit}
Y.~Wang, X.~Pan, S.~Song, H.~Zhang, G.~Huang, and C.~Wu, ``Implicit semantic data augmentation for deep networks,'' in \emph{NeurIPS}, 2019.

\bibitem{van2008visualizing}
L.~Van~der Maaten and G.~Hinton, ``Visualizing data using t-sne.'' \emph{Journal of machine learning research}, vol.~9, no.~11, 2008.

\bibitem{sun2021part}
W.~Sun, J.~Xie, J.~Qiu, and Z.~Ma, ``Part uncertainty estimation convolutional neural network for person re-identification,'' in \emph{ICIP}.\hskip 1em plus 0.5em minus 0.4em\relax IEEE, 2021.

\bibitem{zheng2021exploiting}
K.~Zheng, C.~Lan, W.~Zeng, Z.~Zhang, and Z.-J. Zha, ``Exploiting sample uncertainty for domain adaptive person re-identification,'' in \emph{AAAI}, 2021.

\bibitem{jin2020uncertainty}
X.~Jin, C.~Lan, W.~Zeng, and Z.~Chen, ``Uncertainty-aware multi-shot knowledge distillation for image-based object re-identification,'' in \emph{AAAI}, 2020.

\bibitem{zhou2023adaptive}
X.~Zhou, Y.~Zhong, Z.~Cheng, F.~Liang, and L.~Ma, ``Adaptive sparse pairwise loss for object re-identification,'' in \emph{CVPR}, 2023.

\bibitem{gu2023msinet}
J.~Gu, K.~Wang, H.~Luo, C.~Chen, W.~Jiang, Y.~Fang, S.~Zhang, Y.~You, and J.~Zhao, ``Msinet: Twins contrastive search of multi-scale interaction for object reid,'' in \emph{CVPR}, 2023.

\bibitem{liu2023learning}
F.~Liu, M.~Kim, Z.~Gu, A.~Jain, and X.~Liu, ``Learning clothing and pose invariant 3d shape representation for long-term person re-identification,'' in \emph{ICCV}, 2023, pp. 19\,617--19\,626.

\bibitem{liu2023clothes}
Y.~Liu, H.~Ge, Z.~Wang, Y.~Hou, and M.~Zhao, ``Clothes-changing person re-identification via universal framework with association and forgetting learning,'' \emph{IEEE Transactions on Multimedia}, 2023.

\bibitem{chen2021deep}
J.~Chen, W.-S. Zheng, Q.~Yang, J.~Meng, R.~Hong, and Q.~Tian, ``Deep shape-aware person re-identification for overcoming moderate clothing changes,'' \emph{IEEE Transactions on Multimedia}, vol.~24, pp. 4285--4300, 2021.

\bibitem{cui2023dcr}
Z.~Cui, J.~Zhou, Y.~Peng, S.~Zhang, and Y.~Wang, ``Dcr-reid: Deep component reconstruction for cloth-changing person re-identification,'' \emph{IEEE Transactions on Circuits and Systems for Video Technology}, 2023.

\bibitem{guo2023semantic}
P.~Guo, H.~Liu, J.~Wu, G.~Wang, and T.~Wang, ``Semantic-aware consistency network for cloth-changing person re-identification,'' in \emph{Proceedings of the 31st ACM International Conference on Multimedia}, 2023, pp. 8730--8739.

\bibitem{huang2024meta}
Y.~Huang, Q.~Wu, Z.~Zhang, C.~Shan, Y.~Zhong, and L.~Wang, ``Meta clothing status calibration for long-term person re-identification,'' \emph{IEEE Transactions on Image Processing}, 2024.

\bibitem{li2022visible}
Y.~Li, T.~Zhang, X.~Liu, Q.~Tian, Y.~Zhang, and F.~Wu, ``Visible-infrared person re-identification with modality-specific memory network,'' \emph{IEEE Transactions on Image Processing}, vol.~31, pp. 7165--7178, 2022.

\bibitem{wu2023learning}
L.~Y. Wu, L.~Liu, Y.~Wang, Z.~Zhang, F.~Boussaid, M.~Bennamoun, and X.~Xie, ``Learning resolution-adaptive representations for cross-resolution person re-identification,'' \emph{IEEE Transactions on Image Processing}, 2023.

\bibitem{wu2022pseudo}
L.~Wu, D.~Liu, W.~Zhang, D.~Chen, Z.~Ge, F.~Boussaid, M.~Bennamoun, and J.~Shen, ``Pseudo-pair based self-similarity learning for unsupervised person re-identification,'' \emph{IEEE Transactions on Image Processing}, vol.~31, pp. 4803--4816, 2022.

\bibitem{yang2022uncertainty}
W.~Yang, T.~Zhang, Y.~Zhang, and F.~Wu, ``Uncertainty guided collaborative training for weakly supervised and unsupervised temporal action localization,'' \emph{IEEE Transactions on Pattern Analysis and Machine Intelligence}, vol.~45, no.~4, pp. 5252--5267, 2022.

\bibitem{zhao2018understanding}
J.~Zhao, J.~Li, Y.~Cheng, T.~Sim, S.~Yan, and J.~Feng, ``Understanding humans in crowded scenes: Deep nested adversarial learning and a new benchmark for multi-human parsing,'' in \emph{Proceedings of the 26th ACM international conference on Multimedia}, 2018, pp. 792--800.

\bibitem{wang2021face}
Q.~Wang, P.~Zhang, H.~Xiong, and J.~Zhao, ``Face. evolve: A high-performance face recognition library,'' \emph{arXiv preprint arXiv:2107.08621}, 2021.

\bibitem{zhao2018towards}
J.~Zhao, Y.~Cheng, Y.~Xu, L.~Xiong, J.~Li, F.~Zhao, K.~Jayashree, S.~Pranata, S.~Shen, J.~Xing \emph{et~al.}, ``Towards pose invariant face recognition in the wild,'' in \emph{Proceedings of the IEEE conference on computer vision and pattern recognition}, 2018, pp. 2207--2216.

\bibitem{zhao20183d}
J.~Zhao, L.~Xiong, Y.~Cheng, Y.~Cheng, J.~Li, L.~Zhou, Y.~Xu, J.~Karlekar, S.~Pranata, S.~Shen \emph{et~al.}, ``3d-aided deep pose-invariant face recognition.'' in \emph{IJCAI}, vol.~2, no.~3, 2018, p.~11.

\bibitem{zhao2020towards}
J.~Zhao, S.~Yan, and J.~Feng, ``Towards age-invariant face recognition,'' \emph{IEEE Transactions on Pattern Analysis and Machine Intelligence}, vol.~44, no.~1, pp. 474--487, 2020.

\bibitem{deng2020retinaface}
J.~Deng, J.~Guo, E.~Ververas, I.~Kotsia, and S.~Zafeiriou, ``Retinaface: Single-shot multi-level face localisation in the wild,'' in \emph{Proceedings of the IEEE/CVF conference on computer vision and pattern recognition}, 2020, pp. 5203--5212.

\bibitem{zhang2016joint}
K.~Zhang, Z.~Zhang, Z.~Li, and Y.~Qiao, ``Joint face detection and alignment using multitask cascaded convolutional networks,'' \emph{IEEE signal processing letters}, vol.~23, no.~10, pp. 1499--1503, 2016.

\bibitem{yu2020cocas}
S.~Yu, S.~Li, D.~Chen, R.~Zhao, J.~Yan, and Y.~Qiao, ``Cocas: A large-scale clothes changing person dataset for re-identification,'' in \emph{Proceedings of the IEEE/CVF conference on computer vision and pattern recognition}, 2020, pp. 3400--3409.

\bibitem{wu2022identity}
J.~Wu, H.~Liu, W.~Shi, H.~Tang, and J.~Guo, ``Identity-sensitive knowledge propagation for cloth-changing person re-identification,'' in \emph{2022 IEEE International Conference on Image Processing (ICIP)}.\hskip 1em plus 0.5em minus 0.4em\relax IEEE, 2022, pp. 1016--1020.

\bibitem{ji2022asymmetric}
Z.~Ji, J.~Hu, D.~Liu, L.~Y. Wu, and Y.~Zhao, ``Asymmetric cross-scale alignment for text-based person search,'' \emph{IEEE Transactions on Multimedia}, vol.~25, pp. 7699--7709, 2022.

\end{thebibliography}
}

\vspace{-10mm}

\begin{IEEEbiography}
[{\includegraphics[width=1in,height=1.25in,clip,keepaspectratio]{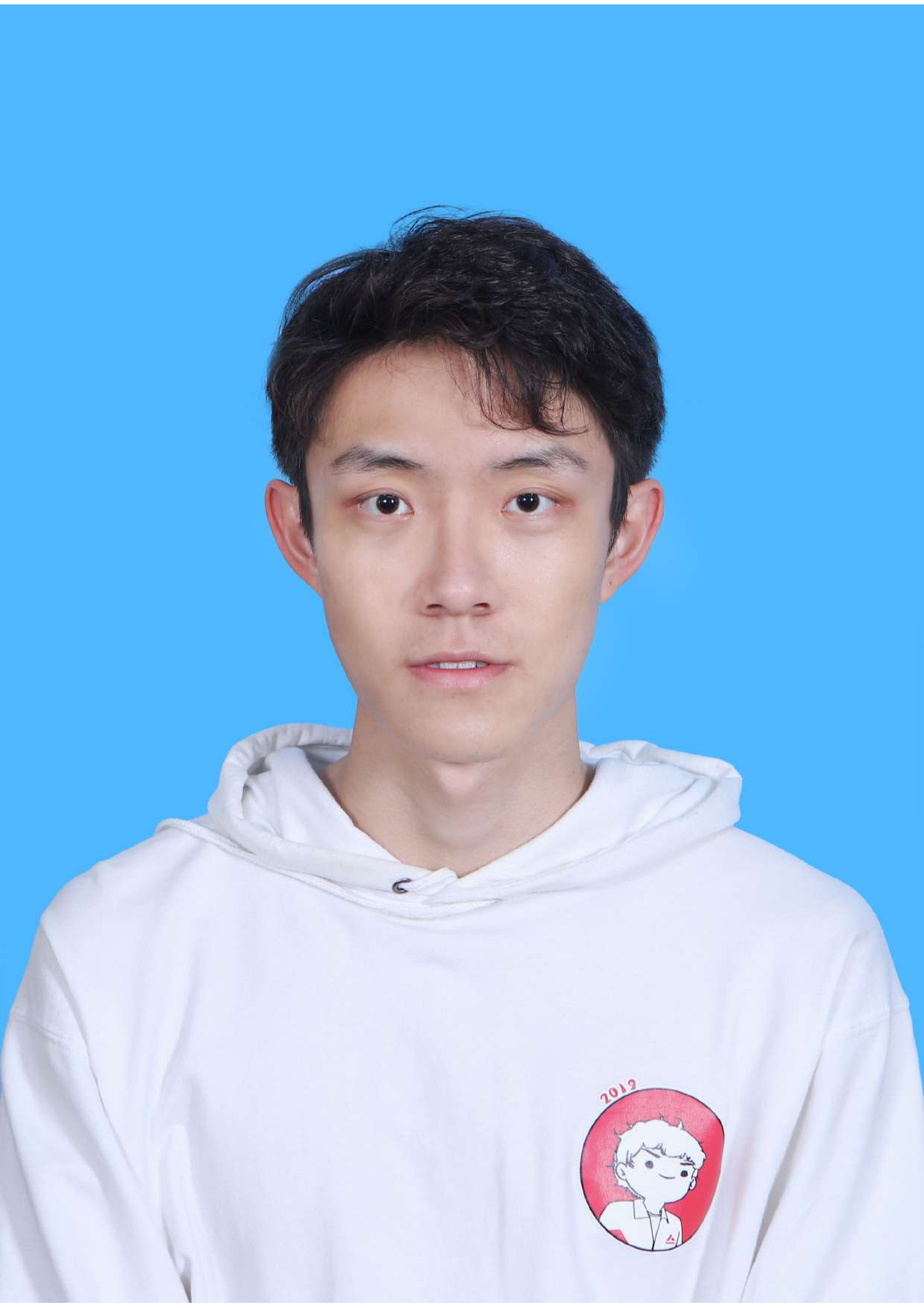}}] {Jiahe Zhao}
received the BS degree in Tsinghua University, Beijing, China in 2022. He is currently pursuing the M.S. degree with the Institute of Computing Technology, Chinese Academy of Sciences, Beijing, China. His research interests are in computer vision and machine learning. He specially focuses on person re-identification and multi-modal large language models.
\end{IEEEbiography}

\vspace{-10mm}

\begin{IEEEbiography}
[{\includegraphics[width=1in,height=1.25in,clip,keepaspectratio]{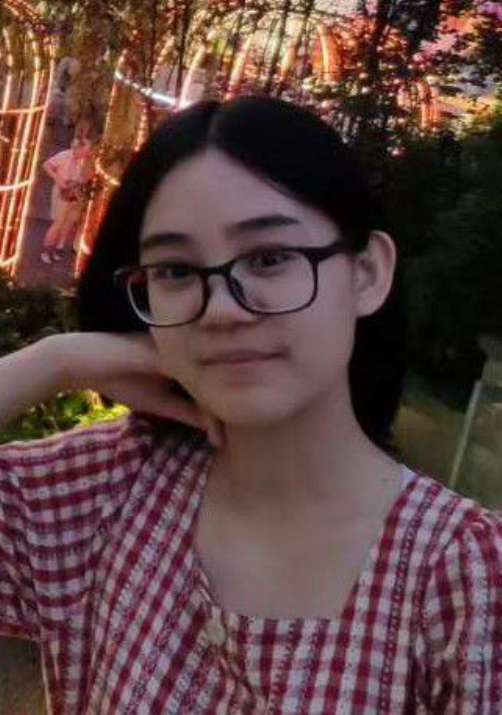}}]{Ruibing Hou}
received the BS degree in Northwestern Polytechnical University, Xi’an, China, in 2016. She received PhD degree in computer science from the Institute of Computing Technology, Chinese Academy of Sciences, Beijing, China, in 2002. She is currently a postdoctorial researcher with the Institute of Computing Technology, Chinese Academy of Sciences. Her research interests are in machine learning and computer vision. She specially focuses on person re-identification, long-tailed learning and few-shot learning. 
\end{IEEEbiography}

\vspace{-10mm}

\begin{IEEEbiography}
[{\includegraphics[width=1in,height=1.25in,clip,keepaspectratio]{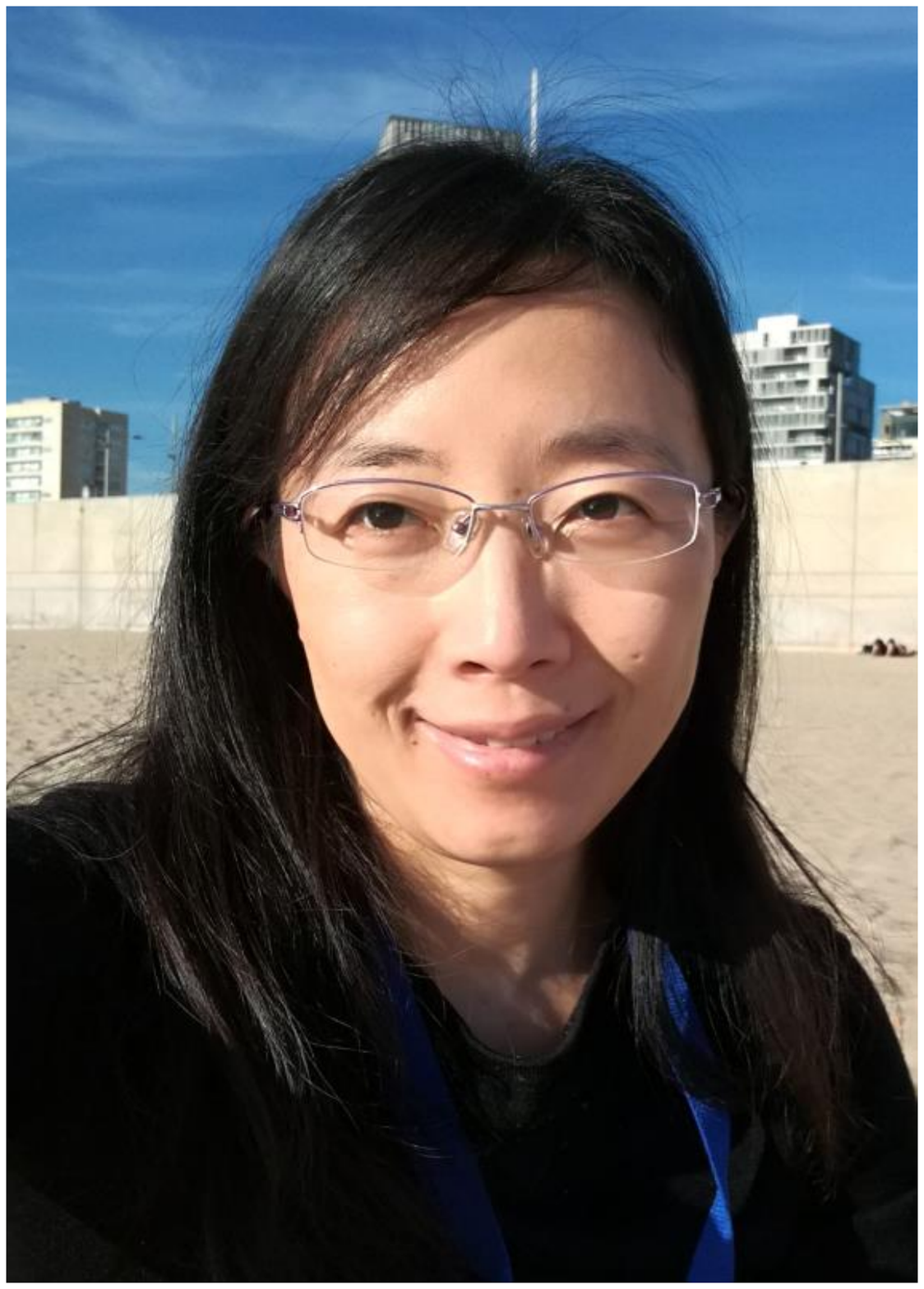}}]{Hong Chang}
received the Bachelor’s degree from Hebei University of Technology, Tianjin,
China, in 1998; the M.S. degree from Tianjin University, Tianjin, in 2001; and the Ph.D. degree from Hong Kong University of Science and Technology, Kowloon, Hong Kong, in 2006, all in computer science. She was a Research Scientist with Xerox Research Centre Europe. She is currently a Researcher with the Institute of Computing Technology, Chinese Academy of Sciences, Beijing, China. Her main research interests include algorithms and models in machine learning, and their applications in pattern recognition and computer vision.
\end{IEEEbiography}

\vspace{-10mm}

\begin{IEEEbiography}
[{\includegraphics[width=1in,height=1.25in,clip,keepaspectratio]{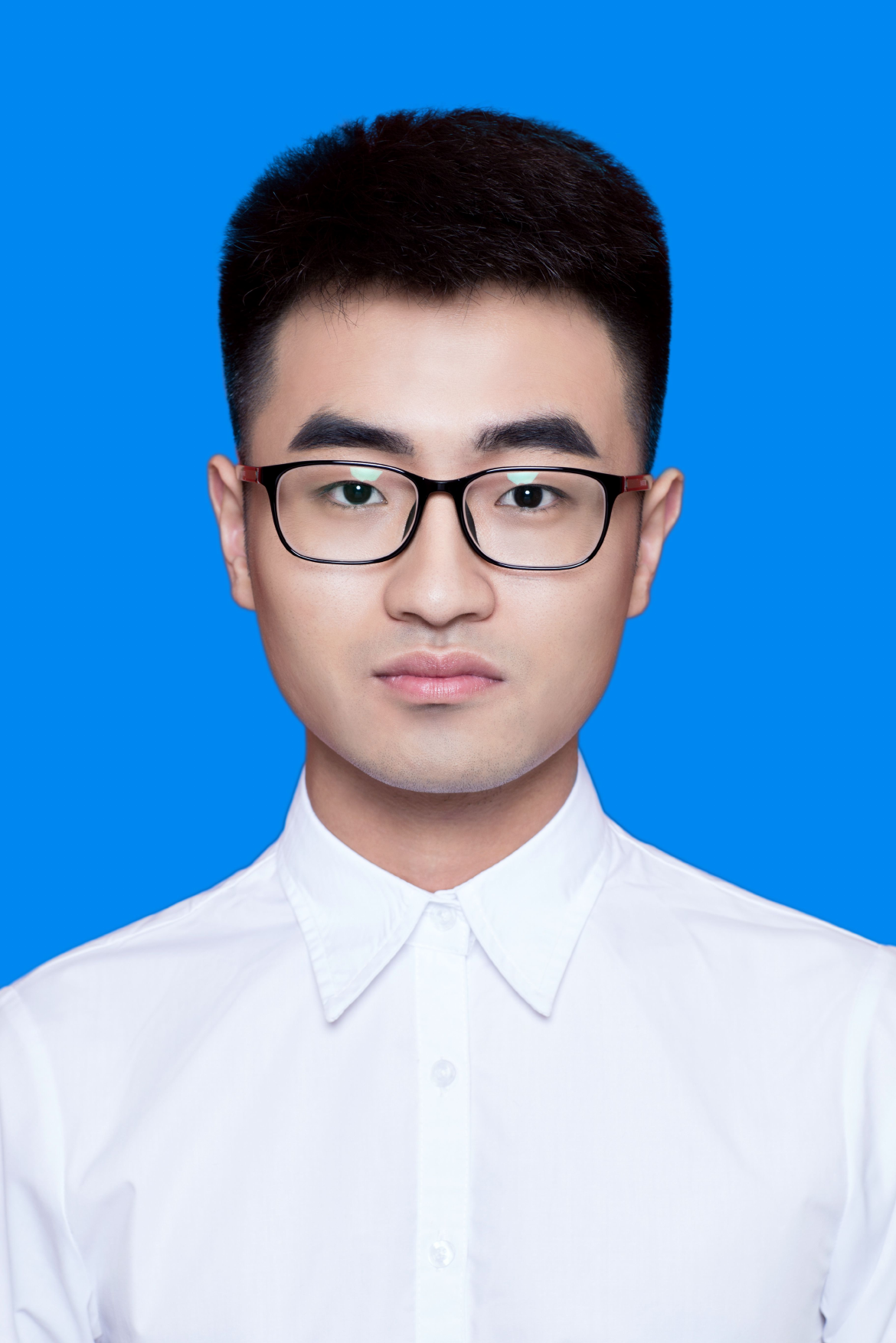}}]{Xinqian Gu}
received the BS degree in software engineering from Chongqing University in 2017. He received PhD degree in computer science from the Institute of Computing Technology, Chinese Academy of Sciences in 2022. His research interests are in computer vision, pattern recognition, and machine learning. He especially focuses on person re-identification, video analytics and the related research topics.
\end{IEEEbiography}

\vspace{-10mm}

\begin{IEEEbiography}
[{\includegraphics[width=1in,height=1.25in,clip,keepaspectratio]{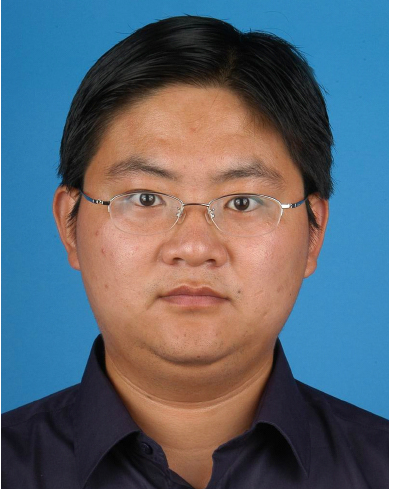}}]{Bingpeng Ma}
received the BS degree in mechanics, in 1998 and the MS degree in mathematics, in 2003 from the Huazhong University of Science and Technology, respectively. He received the PhD degree in computer science from the Institute of Computing Technology, Chinese Academy of Sciences, P.R. China, in 2009. He was a post-doctorial researcher with the University of Caen, France, from 2011 to 2012. He joined the School of Computer Science and Technology, University of Chinese Academy of Sciences, Beijing, in March 2013 and now he is a professor.  His research interests cover computer vision, pattern recognition, and machine learning. He especially focuses on person re-identification, face recognition, and the related research topics.
\end{IEEEbiography}

\vspace{-10mm}

\begin{IEEEbiography}
[{\includegraphics[width=1in,height=1.25in,clip,keepaspectratio]{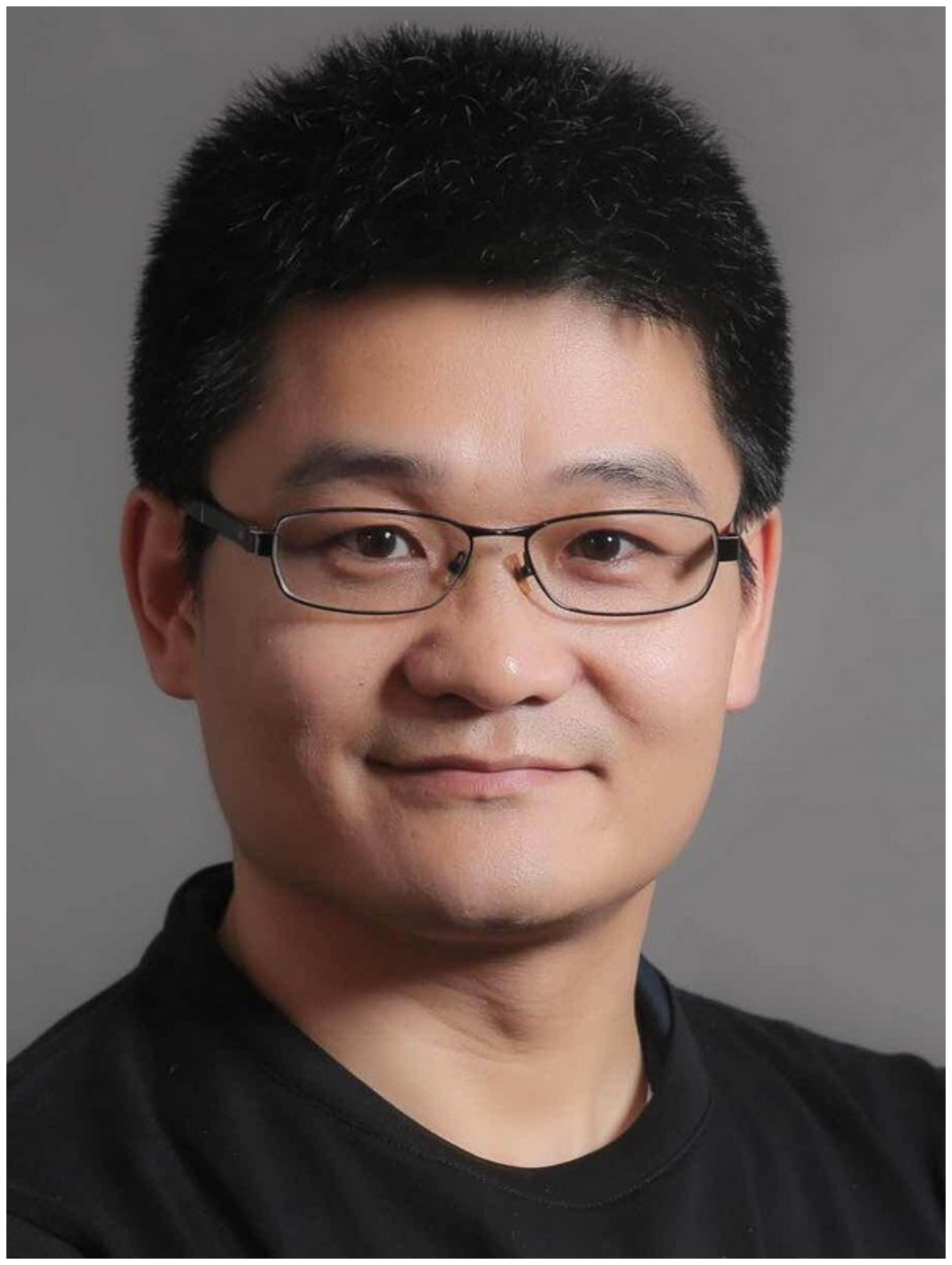}}]{Shiguang Shan}
(M’04-SM’15) received Ph.D. degree in computer science from the Institute of Computing Technology (ICT), Chinese Academy of Sciences (CAS), Beijing, China, in 2004. He has been a full Professor of this institute since 2010 and now the deputy director of CAS Key Lab of Intelligent Information Processing. His research interests cover computer vision, pattern recognition, and machine learning. He has published more than 300 papers, with totally more than 20,000 Google scholar citations. He served as Area Chairs for many international conferences including CVPR, ICCV, AAAI, IJCAI, ACCV, ICPR, FG, etc. And he was/is Associate Editors of several journals including IEEE T-IP, Neurocomputing, CVIU, and PRL. He was a recipient of the China’s State Natural Science Award in 2015, and the China’s State S\&T Progress Award in 2005 for his research work.
\end{IEEEbiography}

\vspace{-10mm}

\begin{IEEEbiography}
[{\includegraphics[width=1in,height=1.25in,clip,keepaspectratio]{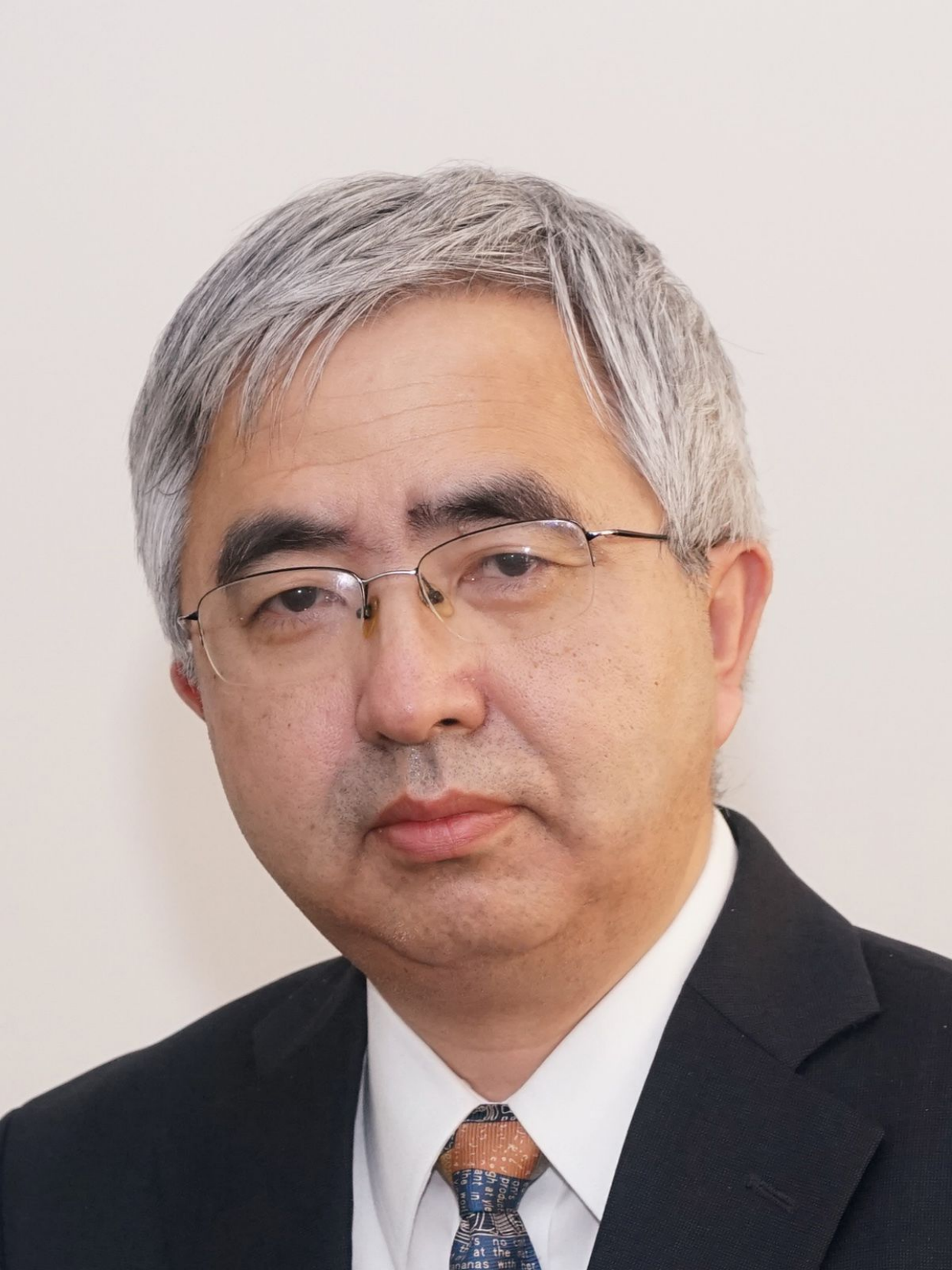}}]{Xilin Chen}
is a professor with the Institute of Computing Technology, Chinese Academy of Sciences (CAS). He has authored one book and more than 300 papers in refereed journals and proceedings in the areas of computer vision, pattern recognition, image processing, and multimodal interfaces. He is currently an information sciences editorial board member of Fundamental Research, an editorial board member of Research, a senior editor of the Journal of Visual Communication and Image Representation, and an associate editor-in-chief of the Chinese Journal of Computers, and Chinese Journal of Pattern Recognition and Artificial Intelligence. He served as an organizing committee member for multiple conferences, including general co-chair of FG13 / FG18, program co-chair of ICMI 2010. He is a fellow of the ACM, IEEE, IAPR, and CCF.
\end{IEEEbiography}

\end{document}